\documentclass[11pt]{article}

\usepackage[final]{acl}

\usepackage{times}
\usepackage{latexsym}

\usepackage[T1]{fontenc}
\usepackage[utf8]{inputenc}

\usepackage{microtype}

\usepackage{inconsolata}

\usepackage{graphicx}

\usepackage{multirow} % for multirow cells in tables
\usepackage{enumitem}
\usepackage{colortbl} % for \arrayrulecolor
\usepackage{microtype}
\usepackage{booktabs, makecell}
\usepackage{amsmath}
\usepackage{amssymb}
\usepackage{mathtools}
\usepackage{amsthm}
\usepackage[table]{xcolor}
\usepackage[most]{tcolorbox}
\usepackage{fontawesome5}
\title{
\texorpdfstring
  {Can LLMs Predict Their Own Failures?\\Self-Awareness via Internal Circuits}
  {Can LLMs Predict Their Own Failures? Self-Awareness via Internal Circuits}
}

\author{Amirhosein Ghasemabadi \\
    University of Alberta, Canada\\
  \texttt{ghasemab@ualberta.ca} \\\And
  Di Niu \\
    University of Alberta, Canada\\
  \texttt{dniu@ualberta.ca} \\}
\begin{document}
\maketitle
\begin{abstract}
Large language models (LLMs) generate fluent and complex outputs but often fail to recognize their own mistakes and hallucinations. Existing approaches typically rely on external judges, multi-sample consistency, or text-based self-critique, which incur additional compute or correlate weakly with true correctness.
We ask: can LLMs predict their own failures by inspecting internal states during inference?
We introduce Gnosis, a lightweight self-awareness mechanism that enables frozen LLMs to perform intrinsic self-verification by decoding signals from hidden states and attention patterns. Gnosis passively observes internal traces, compresses them into fixed-budget descriptors, and predicts correctness with negligible inference cost, adding only ~5M parameters and operating independently of sequence length.
Across math reasoning, open-domain question answering, and academic knowledge benchmarks, and over frozen backbones ranging from 1.7B to 20B parameters, Gnosis consistently outperforms strong internal baselines and large external judges in both accuracy and calibration. Moreover, it generalizes zero-shot to partial generations, enabling early detection of failing trajectories and compute-aware control. These results show that reliable correctness cues are intrinsic to generation process and can be extracted efficiently without external supervision. 

Code and models: \href{https://github.com/Amirhosein-gh98/Gnosis}{\faGithub\ Gnosis Github}.

%Large language models (LLMs) are increasingly capable at open-ended generation and multi-step reasoning, yet they remain unreliable at assessing when their own answers are incorrect. We introduce \textbf{Gnosis}, a lightweight self-awareness mechanism that retrofits frozen LLMs with introspection by \emph{reading} internal traces rather than invoking external judges. Gnosis is intentionally compact, adding only $\sim$5M parameters, roughly $1000\times$ smaller than 8B reward models. We evaluate Gnosis in three practical regimes: (i) self-judgment on each backbone's own generations, (ii) sibling-model judgment where a small head serves as a lightweight reward model for larger family members, and (iii) early correctness prediction on partial completions for compute-aware control. Across math reasoning, open-domain QA, and academic knowledge reasoning, and across five frozen backbones (Qwen3 1.7B/4B/8B variants and gpt-oss-20B), Gnosis consistently outperforms strong internal baselines and large external judges while enabling earlier detection of failing trajectories.
\end{abstract}

\section{Introduction}

%Large language models (LLMs) excel at open-ended generation and multi-step reasoning, but remain unreliable at judging their \emph{own} outputs. They routinely miss their own mistakes and hallucinations, and often assign high confidence to incorrect answers. This gap between strong generation and weak self-verification limits reliability, safety, and compute-efficient deployment.
Large language models (LLMs) have achieved remarkable performance in open-ended generation and multi-step reasoning, yet they remain unreliable at assessing the correctness of their own outputs\cite{kalai2025language, huang2025survey}. They frequently produce confident but incorrect answers, failing to detect reasoning errors or hallucinations even when such failures are evident to external evaluators\cite{kirichenko2025abstentionbench, kamoi2024evaluating}. This gap between strong generation and weak self-verification limits the reliability, safety, and efficiency of LLM deployment, particularly in settings that require long-horizon reasoning or compute-aware control. A fundamental open question is whether LLMs can anticipate their own failures by examining the internal dynamics that govern their generation process.

Prior work on LLM self-evaluation and hallucination detection largely follows three paradigms. \textbf{Text-based self-critique and confidence estimation}~\citep{kadavath2022language,ulmer2024calibrating, huang2025efficient} infer correctness from generated text or token probabilities, often tracking linguistic fluency rather than reasoning validity and degrading on long or compositional tasks. 
\textbf{Multi-sample consistency} methods~\citep{sriramanan2024llmcheck, pawitan2025confidence} estimate confidence from agreement across multiple generations, improving robustness at the cost of inference that scales linearly with the number of samples. 
\textbf{External judges and reward models} ~\citep{stiennon2020learning, ouyang2022training, zheng2024processbench, wang2024helpsteer2,liu2025skywork} train large auxiliary models to evaluate responses, providing strong signals but requiring costly supervision, additional decoding passes, and substantial inference overhead. Despite their differences, these approaches rely on signals external to the model’s own internal dynamics, leaving open whether correctness can be predicted directly from the generation process itself.

\begin{figure*}[t]
    \centering
    \includegraphics[width=1\textwidth]{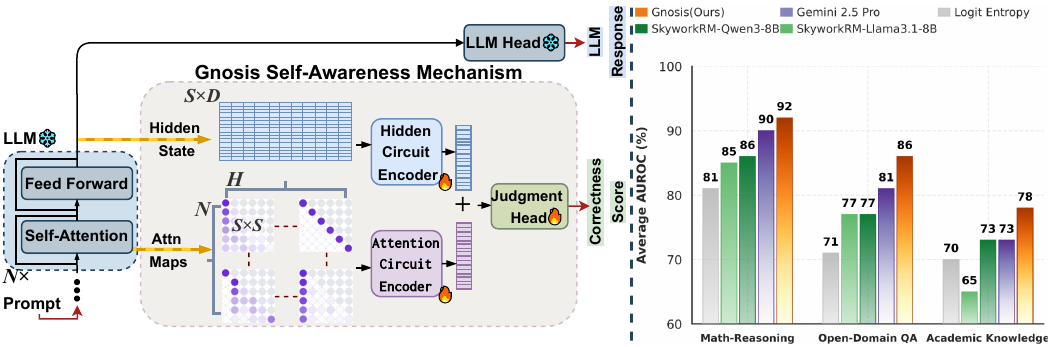}
    \vspace{-2em}
    \caption{Overview of our \textbf{Gnosis} self-awareness mechanism and its performance. 
    \textbf{Left:} Gnosis taps hidden states and attention maps from a frozen LLM, learns to compress them into hidden/attention descriptors, and predicts a scalar correctness (hallucination) score with only $\sim$5 million extra parameters and essentially zero added inference cost. 
    \textbf{Right:} Gnosis outperforms 8B Skywork reward models and a Gemini~2.5 Pro judge in AUROC on Math-Reasoning (AMC12 + AIME24/25 + HMMT Feb~2025), Open-Domain QA (TriviaQA), and Academic Knowledge (MMLU-Pro); scores are averaged over the frozen backbones listed in Table~\ref{table1}.}

    \label{fig:main_fig}
\end{figure*}

In this paper, we demonstrate that large language models can reliably predict their own failures by leveraging signals intrinsic to the generation process. We introduce Gnosis, a lightweight self-awareness mechanism that endows frozen LLMs with intrinsic self-verification, eliminating the need for external judges.
% , multi-sample decoding, or task-specific self-critique. 
By extracting reliability cues directly from model-internal dynamics during inference, Gnosis produces accurate and well-calibrated correctness estimates with negligible computational overhead. This intrinsic capability enables early detection of failing reasoning trajectories, efficient scaling across model sizes and domains, and practical deployment of compute-aware and reliability-critical language systems. Our main contributions are:

% Moving beyond external judges, we introduce Gnosis, a lightweight mechanism that enables frozen LLMs to introspect internal generation traces. This provides a form of introspective awareness about the model’s likely success on a given task. By decoding intrinsic signals from hidden states and attention and compressing them into fixed-budget descriptors, Gnosis predicts correctness with negligible compute and \textbf{no additional decoding cost}.
% Despite being orders of magnitude smaller than external verifiers, Gnosis achieves state-of-the-art performance, consistently outperforming billion-parameter Reward Models and Large proprietary models like Gemini 2.5 Pro\cite{comanici2025gemini}. Furthermore, its design enables the \textbf{early detection} of failing trajectories via prefix scanning, supporting compute-aware control policies without altering the backbone. Across 1B--20B Dense and MoE backbones and three domains spanning \textbf{math reasoning}, \textbf{open-domain QA}, and \textbf{academic knowledge}, Gnosis consistently surpasses strong internal baselines and matches or exceeds the tested external verifiers, establishing a robust new baseline for efficient self-evaluation.

\vspace{-0.5em}
\begin{itemize}
    \item \textbf{Intrinsic, Trajectory-Level Self-Awareness.} We introduce Gnosis, a lightweight mechanism that enables frozen LLMs to predict the correctness of their own generations by decoding signals intrinsic to the inference process. Unlike prior internal-signal methods that rely on statistical features \cite{geng2023survey, wang2025latentcoe, zhang2025hiddenstates} or single-token indicators\cite{ zhang2025reasoning}, Gnosis leverages the full spatiotemporal structure of internal dynamics across an entire generation trajectory.
    % \vspace{-0.5em}
    \item \textbf{Dual-Stream Introspection from Hidden States and Attention.} Gnosis jointly models hidden-state evolution and attention-routing patterns through a compact, fixed-budget architecture that operates independently of sequence length, extracting rich reliability cues with negligible inference overhead.
    % \vspace{-0.5em}
    \item \textbf{Cross-Scale Transfer and Early Failure Detection.} Gnosis generalizes beyond self-judgment: a head trained on a small backbone model transfers zero-shot to larger variants, and predicts failures reliably from partial reasoning and generations, enabling early termination and compute-aware control.
    % \vspace{-0.5em}
    % \item \textbf{State-of-the-Art Performance at Minimal Scale.} With only ~5M additional parameters, Gnosis is orders of magnitude smaller than external verifiers, yet outperforms billion-parameter reward models and proprietary judges such as Gemini 2.5 Pro across math reasoning, open-domain QA, and academic knowledge benchmarks. It establishes a \textcolor{red}{robust new baseline} across frozen 1B–20B backbones while maintaining near-constant inference latency.
    % \item \textbf{State-of-the-Art Performance at Minimal Scale.} With only $\sim$5M additional parameters, Gnosis is orders of magnitude smaller than external verifiers, yet outperforms billion-parameter reward models and proprietary judges (e.g., Gemini 2.5 Pro) across math reasoning, open-domain QA, and academic knowledge benchmarks. It works reliably across diverse frozen backbones of different architectures and sizes, with negligible latency overhead.
    \item \textbf{State-of-the-Art Performance at Minimal Scale.} With only $\sim$5M added parameters, Gnosis is orders of magnitude smaller than external verifiers yet outperforms billion-parameter reward models and proprietary judges on math reasoning, open-domain QA, and academic benchmarks. It works reliably across diverse frozen backbones with negligible latency overhead.
    % \vspace{-1em}
\end{itemize}

\section{Related Work}
\vspace{-.5em}
Methods for assessing LLM correctness and hallucination risk largely fall into four families: Text-based confidence \& self-critique, Internal signal-based indicators and linear probes, external reward models and judge LLMs, and multi-sample self-consistency methods.

\textbf{External Reward Models and Judge LLMs.}
External verifiers train separate models to score response quality, factuality, or step-wise correctness. Outcome and Process Reward Models (ORM/PRM) are widely used for ranking, hallucination detection, and guiding test-time search~\citep{stiennon2020learning,ouyang2022training,zheng2024processbench,wang2023math,zhang2025lessons}. Recent systems emphasize large, carefully curated datasets over architectural novelty: HelpSteer2 combines Likert ratings, pairwise preferences, and extrapolation to sharpen discrimination~\citep{wang2024helpsteer2}, while Skywork-Reward-V2 scales human--AI curation to tens of millions of preference pairs and leads on RewardBench-style suites~\citep{liu2025skywork}. These models provide strong signals but incur substantial annotation cost and add inference latency and deployment overhead by requiring a large auxiliary model at serving time.

\textbf{ Text-Based Confidence \& Self-Critique.}
Text-based approaches aim to estimate correctness from the generated text and token probabilities. Training-free indicators use entropy or max probability as uncertainty proxies but struggle with confident hallucinations and out-of-distribution shifts~\citep{geng2023survey,sriramanan2024llm,pawitan2025confidence}. Prompt-based calibration elicits verbalized confidence or self-critique, improving ECE but often tracking stylistic fluency more than reasoning validity and requiring extra passes~\citep{kadavath2022language,ulmer2024calibrating}. Generative and distillation-based calibrators predict correctness in a single forward pass, e.g., APRICOT trains a calibrator LLM~\citep{ulmer2024calibrating}, and Self-Calibration distills self-consistency signals to enable early stopping and confidence-weighted sampling~\citep{huang2025efficient}. These methods reduce dependence on external judges but may require full-model fine-tuning, add training cost, and remain brittle across domains and sequence lengths.

\textbf{Internal signal-based indicators and linear probes.}
Glass-box signals exploit logits, hidden states, and attention routing. Prior work shows hidden activations diverge between correct and hallucinated outputs~\citep{duan2024hallucination}, with factuality cues concentrated in middle/deep layers yet sensitive to domain shift~\citep{zhang2025hiddenstates}. Token-wise hidden-state entropy and information density can outperform perplexity-based failure prediction~\citep{chen2024measureinfo}. Trajectory/spectral views analyze how representations evolve across layers (e.g., Chain-of-Embedding, stability of latent paths) and relate angular/magnitude changes to correctness~\citep{wang2025latentcoe}. Attention statistics provide lightweight reliability cues~\citep{huang2024selfeval}. A complementary line trains simple probes (shallow MLPs) on final-token states~\citep{azaria2023internal, burns2022discovering, zhang2025reasoning}. However, these approaches consistently yield low accuracy across diverse benchmarks. By relying on fragile heuristics or single-token snapshots, they miss the generation's full spatiotemporal structure, resulting in performance that falls far short of Gnosis.

\textbf{Multi-Sample Self-Consistency and Test-Time Scaling.}
Multi-sample self-consistency infers confidence from agreement across sampled rationales, boosting robustness but incurring inference cost that scales with the number of samples and often saturating on long, compositional tasks~\citep{sriramanan2024llm}. Recent cost-aware test-time scaling uses internal signals to prune search or adapt compute, reducing dependence on large external verifiers while retaining some benefits of multi-sample reasoning~\citep{huang2025efficient,ghasemabadi2025guided}.

\begin{table*}[t]
\centering
\small
\setlength{\tabcolsep}{2pt}
\renewcommand{\arraystretch}{0.95}
\caption{Correctness/hallucination detection across domains. For each model, columns (left to right) are: AUROC / AUPR-c / AUPR-e / BSS / ECE.}
\begin{tabular}{l|ccccc|ccccc|ccccc|ccccc}
\toprule
\textbf{Method} &
\multicolumn{5}{c|}{\textbf{Qwen3 1.7B-Hybrid}} &
\multicolumn{5}{c|}{\textbf{Qwen3 4B-Thinking}} &
\multicolumn{5}{c|}{\textbf{Qwen3 4B-Instruct}} &
\multicolumn{5}{c}{\textbf{OpenAI gpt-oss-20B}} \\
\midrule
\multicolumn{21}{c}{\small(AUROC $\uparrow$ \;/\; AUPR-c $\uparrow$ \;/\; AUPR-e $\uparrow$ \;/\; BSS $\uparrow$ \;/\; ECE $\downarrow$)} \\
\midrule

% ================== Domain I ==================
\multicolumn{21}{c}{\textbf{Domain I: Math-Reasoning(AMC12 + AIME24/25 + HMMTFeb2025)}} \\
\midrule
Logit Entropy~\citeyearpar{sriramanan2024llmcheck} &
.79 & .73 & .82 & .25 & \textbf{.05} &
.80 & .80 & .77 & .23 & \underline{.12} &
.83 & .79 & \underline{.82} & \underline{.32} & .74 &
.80 & .79 & .81 & \underline{.32} & \underline{.07} \\

Mean Token Prob~\citeyearpar{sriramanan2024llmcheck} &
.78 & .71 & .82 & .23 & \underline{.06} &
.80 & .79 & .76 & .21 & .13 &
.82 & .79 & \underline{.82} & .31 & .08 &
.79 & .78 & .80 & .30 & \underline{.07} \\

Attn Eigenvalue Score~\citeyearpar{sriramanan2024llmcheck} &
.61 & .52 & .63 & -.13 & .17 &
.55 & .60 & .46 & -.28 & .23 &
.72 & .66 & .75 & .11 & \underline{.13} &
.72 & .66 & .75 & .11 & .13 \\

CoE--R~\citeyearpar{wang2025latentcoe} &
.59 & .51 & .63 & -.17 & .20 &
.56 & .60 & .55 & -.26 & .24 &
.66 & .65 & .59 & -.02 & .15 &
.66 & .66 & .60 & -.02 & .15 \\

CoE--C~\citeyearpar{wang2025latentcoe} &
.60 & .53 & .64 & -.14 & .18 &
.53 & .57 & .52 & -.32 & .26 &
.66 & .66 & .59 & -.02 & .14 &
.66 & .66 & .59 & -.02 & .14 \\
\arrayrulecolor{gray!60}\specialrule{0.15pt}{0pt}{1pt}\arrayrulecolor{black}

SkyworkRM-Llama3.1-8B~\citeyearpar{liu2025skywork} &
.88 & .88 & .88 & .38 & .10 &
.87 & .93 & \underline{.80} & .24 & .18 &
.83 & \underline{.90} & .72 & .06 & .22 &
.81 & .80 & .83 & .29 & .10 \\

SkyworkRM-Qwen3-8B~\citeyearpar{liu2025skywork} &
.90 & \underline{.92} & \underline{.89} & .39 & .14 &
.89 & .94 & .77 & .10 & .22 &
.83 & .89 & .75 & -.49 & .40 &
.81 & .79 & \underline{.84} & -.12 & .30 \\

Gemini 2.5 Pro~\citeyearpar{comanici2025gemini} &
\underline{.91} & .88 & .86 & \underline{.50} & .11 &
\underline{.92} & \underline{.97} & .68 & \underline{.46} & .15 &
\underline{.84} & .88 & .66 & .31 & .18 &
\textbf{.92} & \textbf{.98} & .64 & -1.09 & .10 \\
\arrayrulecolor{gray!60}\specialrule{0.15pt}{0pt}{1pt}\arrayrulecolor{black}

\rowcolor{blue!6}
Gnosis(Ours) &
\textbf{.95} & \textbf{.95} & \textbf{.94} & \textbf{.59} & .09 &
\textbf{.96} & \textbf{.98} & \textbf{.91} & \textbf{.65} & \textbf{.05} &
\textbf{.93} & \textbf{.96} & \textbf{.89} & \textbf{.51} & \textbf{.08} &
\underline{.85} & \underline{.86} & \textbf{.86} & \textbf{.38} & \textbf{.04} \\
\midrule

% ================== Domain II ==================
\multicolumn{21}{c}{\textbf{Domain II: Open-Domain QA(TriviaQA)}} \\
\midrule
Logit Entropy &
.64 & .53 & .73 & -.16 & .19 &
.68 & .70 & .63 & .02 & \underline{.13} &
.71 & \underline{.75} & .64 & \textbf{.74} & \underline{.11} &
.79 & .85 & .63 & \underline{.05} & .21 \\

Mean Token Prob &
.63 & .52 & .72 & -.17 & .19 &
.67 & .70 & .63 & .00 & .14 &
.71 & \underline{.75} & .64 & \underline{.71} & \underline{.11} &
.79 & .86 & .62 & \underline{.05} & .20 \\

Attn Eigenvalue Score &
.52 & .40 & .62 & -.39 & .27 &
.57 & .61 & .50 & -.20 & .19 &
.59 & .65 & .51 & -.16 & .19 &
.65 & .81 & .40 & -.30 & .21 \\

CoE--R &
.59 & .44 & .70 & -.25 & .22 &
.53 & .57 & .49 & -.29 & .24 &
.57 & .62 & .48 & -.23 & .19 &
.78 & .86 & .61 & -.03 & .21 \\

CoE--C &
.58 & .42 & .70 & -.28 & .22 &
.59 & .61 & .54 & -.17 & .20 &
.52 & .57 & .44 & -.32 & .22 &
.78 & .87 & .61 & -.03 & .21 \\
\arrayrulecolor{gray!60}\specialrule{0.15pt}{0pt}{1pt}\arrayrulecolor{black}

SkyworkRM-Llama3.1-8B &
.83 & \underline{.74} & .87 & .00 & .25 &
.75 & .74 & .74 & -.13 & .28 &
.69 & .73 & .60 & -.47 & .37 &
.79 & \underline{.88} & .59 & -1.11 & .52 \\

SkyworkRM-Qwen3-8B &
.84 & .73 & .89 & .00 & .23 &
.73 & \textbf{.89} & .43 & -.05 & .17 &
.67 & .71 & .57 & -.82 & .47 &
\underline{.82} & \textbf{.90} & \underline{.64} & -1.54 & .59 \\

Gemini 2.5 Pro &
\textbf{.90} & \textbf{.79} & \underline{.91} & \textbf{.40} & \underline{.11} &
\underline{.84} & \underline{.80} & \underline{.86} & \underline{.33} & .14 &
\underline{.75} & \underline{.75} & \underline{.67} & -.02 & .23 &
.74 & .83 & .54 & -.01 & \underline{.20} \\
\arrayrulecolor{gray!60}\specialrule{0.15pt}{0pt}{1pt}\arrayrulecolor{black}

\rowcolor{blue!6}
Gnosis(Ours) &
\underline{.87} & \textbf{.79} & \textbf{.92} & \underline{.34} & \textbf{.10} &
\textbf{.89} & \textbf{.89} & \textbf{.88} & \textbf{.45} & \textbf{.05} &
\textbf{.86} & \textbf{.87} & \textbf{.84} & .38 & \textbf{.05} &
\textbf{.83} & \textbf{.90} & \textbf{.73} & \textbf{.19} & \textbf{.17} \\
\midrule

% ================== Domain III ==================
\multicolumn{21}{c}{\textbf{Domain III: Academic Knowledge-Reasoning(MMLU-Pro)}} \\
\midrule
Logit Entropy &
.73 & .86 & .49 & -.11 & .20 &
\underline{.74} & \underline{.90} & .41 & -.31 & .25 &
\underline{.70} & .82 & .45 & -.22 & .21 &
.61 & .75 & .39 & -.36 & .23 \\

Mean Token Prob &
.73 & .86 & .49 & -.11 & .19 &
\underline{.74} & .89 & .42 & -.30 & .25 &
\underline{.70} & \underline{.83} & \underline{.48} & -.17 & .23 &
.65 & .78 & .31 & -.24 & .20 \\

Attn Eigenvalue Score &
.61 & .78 & .36 & -.35 & .22 &
.51 & .78 & .24 & -.76 & .29 &
.59 & .80 & .32 & -.47 & .23 &
.52 & .72 & .31 & -.50 & .26 \\

CoE--R &
.55 & .72 & .34 & -.47 & .25 &
.59 & .80 & .32 & -.60 & .30 &
.52 & .72 & .28 & -.64 & .29 &
.61 & .77 & .38 & -.33 & .22 \\

CoE--C &
.55 & .72 & .34 & -.47 & .25 &
.60 & .81 & .32 & -.58 & .28 &
.52 & .73 & .29 & -.62 & .29 &
.60 & .76 & .38 & -.34 & .22 \\
\arrayrulecolor{gray!60}\specialrule{0.15pt}{0pt}{1pt}\arrayrulecolor{black}

SkyworkRM-Llama3.1-8B &
.65 & .79 & .46 & .01 & \textbf{.10} &
.61 & .82 & .37 & -.03 & \underline{.10} &
.61 & .80 & .38 & -.11 & \underline{.15} &
.71 & .82 & .53 & -.39 & .33 \\

SkyworkRM-Qwen3-8B &
\underline{.76} & \underline{.87} & .53 & .01 & .17 &
.73 & .88 & .43 & -.05 & .17 &
.66 & .82 & .45 & -.57 & .35 &
\underline{.75} & \textbf{.86} & \underline{.57} & -1.02 & .50 \\

Gemini 2.5 Pro &
\underline{.76} & .83 & \textbf{.68} & \underline{.12} & .16 &
.70 & .87 & \underline{.49} & \underline{-.01} & .18 &
.67 & \underline{.83} & .37 & -.28 & .24 &
\textbf{.78} & \underline{.84} & \textbf{.70} & \textbf{.22} & \underline{.15} \\
\arrayrulecolor{gray!60}\specialrule{0.15pt}{0pt}{1pt}\arrayrulecolor{black}

\rowcolor{blue!6}
Gnosis(Ours) &
\textbf{.80} & \textbf{.90} & \underline{.56} & \textbf{.15} & \underline{.11} &
\textbf{.82} & \textbf{.93} & \textbf{.55} & \textbf{.21} & \textbf{.05} &
\textbf{.74} & \textbf{.87} & \textbf{.51} & \textbf{.10} & \textbf{.05} &
\underline{.75} & \underline{.84} & .51 & \underline{.07} & \textbf{.06} \\
\bottomrule
\end{tabular}
\label{table1}
\end{table*}

\begin{table}[t]
\centering
\small
\setlength{\tabcolsep}{3pt}
\renewcommand{\arraystretch}{0.95}
\caption{Comparison with an MLP-Prob baseline on \textbf{Qwen3 1.7B}.}
\label{table2}
\begin{tabular}{l|ccc|ccc|ccc}
\toprule
\textbf{Method} &
\multicolumn{3}{c|}{\textbf{Math}} &
\multicolumn{3}{c|}{\textbf{TriviaQA}} &
\multicolumn{3}{c}{\textbf{MMLU-Pro}} \\
\midrule
\multicolumn{10}{c}{\scriptsize(AUROC $\uparrow$, AUPR-c $\uparrow$, ECE $\downarrow$)} \\
\midrule
MLP-Prob~\citeyearpar{zhang2025reasoning} &
.86 & .85 & .19 &
.71 & .58 & .21 &
.69 & .79 & .23 \\
\rowcolor{blue!6}
Gnosis &
\textbf{.95} & \textbf{.95} & \textbf{.09} &
\textbf{.87} & \textbf{.79} & \textbf{.10} &
\textbf{.80} & \textbf{.90} & \textbf{.11} \\
\bottomrule
\end{tabular}
\end{table}

\newcommand{\smallerthansmall}{\fontsize{8}{10.2}\selectfont}
\begin{table}[t]
\centering
\smallerthansmall
\setlength{\tabcolsep}{2pt}
\renewcommand{\arraystretch}{0.85}
\caption{Sibling-model judgment across domains. Each triplet of columns shows AUROC / AUPR-c / ECE. 
\textbf{Gnosis-SelfJudge}: Gnosis head trained on each backbone and judging its own generations. 
\textbf{Gnosis-RM}: a single Gnosis head trained on \textbf{Qwen3 1.7B-Hybrid} and used as a reward model for the other models.}
\label{tab:sibling-judging}
\begin{tabular}{l|ccc|ccc|ccc}
\toprule
\textbf{Judge / Model} &
\multicolumn{3}{c|}{\begin{tabular}[c]{@{}c@{}}\textbf{Qwen3}\\ \textbf{1.7B-Hybrid}\end{tabular}} &
\multicolumn{3}{c|}{\begin{tabular}[c]{@{}c@{}}\textbf{Qwen3}\\ \textbf{4B-Thinking}\end{tabular}} &
\multicolumn{3}{c}{\begin{tabular}[c]{@{}c@{}}\textbf{Qwen3}\\ \textbf{8B-Hybrid}\end{tabular}} \\
\midrule
\multicolumn{10}{c}{\scriptsize(AUROC $\uparrow$, AUPR-c $\uparrow$, ECE $\downarrow$)} \\
\midrule

% ================== Domain I ==================
\multicolumn{10}{c}{\textbf{Domain I: Math(AMC12 + AIME24/25 + HMMTFeb2025)}} \\
\midrule
SkyworkRM-Qwen3-8B &
.90 & .92 & .14 &
.89 & .94 & .22 &
.86 & .95 & .23 \\
\arrayrulecolor{gray!60}\specialrule{0.15pt}{0pt}{1pt}\arrayrulecolor{black}
Gnosis-SelfJudge &
\multirow{2}{*}{\textbf{.95}} &
\multirow{2}{*}{\textbf{.95}} &
\multirow{2}{*}{\textbf{.09}} &
\textbf{.96} & \textbf{.98} & \textbf{.05} &
\textbf{.97} & \underline{.97} & \underline{.08} \\
Gnosis-Qwen1.7B as RM &
 &  &  &
\underline{.93} & \underline{.97} & \underline{.18} &
\textbf{.97} & \textbf{.99} & \textbf{.07} \\
\midrule

% ================== Domain II ==================
\multicolumn{10}{c}{\textbf{Domain II: Open-Domain QA(TriviaQA)}} \\
\midrule
SkyworkRM-Qwen3-8B &
.84 & .73 & .23 &
.73 & \textbf{.89} & .17 &
.72 & .78 & .32 \\
\arrayrulecolor{gray!60}\specialrule{0.15pt}{0pt}{1pt}\arrayrulecolor{black}
Gnosis-SelfJudge &
\multirow{2}{*}{\textbf{.87}} &
\multirow{2}{*}{\textbf{.79}} &
\multirow{2}{*}{\textbf{.10}} &
\textbf{.89} & \textbf{.89} & \underline{.05} &
\textbf{.86} & \textbf{.90} & \textbf{.09} \\
Gnosis-Qwen1.7B as RM &
 &  &  &
\underline{.86} & \underline{.86} & \textbf{.04} &
\underline{.84} & \underline{.88} & \underline{.12} \\
\midrule

% ================== Domain III ==================
\multicolumn{10}{c}{\textbf{Domain III: Academic Knowledge-Reasoning(MMLU-Pro)}} \\
\midrule
SkyworkRM-Qwen3-8B &
.76 & .87 & .17 &
.73 & \underline{.88} & .17 &
.73 & \underline{.89} & .19 \\
\arrayrulecolor{gray!60}\specialrule{0.15pt}{0pt}{1pt}\arrayrulecolor{black}
Gnosis-SelfJudge &
\multirow{2}{*}{\textbf{.80}} &
\multirow{2}{*}{\textbf{.90}} &
\multirow{2}{*}{\textbf{.11}} &
\textbf{.82} & \textbf{.93} & \textbf{.05} &
\textbf{.83} & \textbf{.94} & \textbf{.07} \\
Gnosis-Qwen1.7B as RM &
 &  &  &
\underline{.81} & \textbf{.93} & \underline{.16} &
\textbf{.83} & \textbf{.94} & \underline{.15} \\
\bottomrule
\end{tabular}
\end{table}

% \begin{table}[t]
% \centering
% \scriptsize
% \setlength{\tabcolsep}{2.5pt}
% \renewcommand{\arraystretch}{0.9}
% \caption{\textbf{Math} on \textbf{Qwen3 1.7B-Hybrid} with two max response lengths.
% Values are reported in the order: AUROC / AUPR-c / ECE. Latency is end-to-end per example.}
% \label{tab:len_sweep_math_qwen17b_clean}
% \begin{tabular}{l|cccc|cccc}
% \toprule
% \textbf{Method} &
% \multicolumn{8}{@{}l@{}}{%
% \begin{tabular}[c]{@{}lcc@{}}
% \, Max Length: & \textbf{12k} & \textbf{\quad\quad\quad\quad\quad\quad\quad\quad24k}
% \end{tabular}} \\
% \cmidrule(lr){2-5}\cmidrule(lr){6-9}
% & \textbf{Latency(ms)} & \multicolumn{3}{c|}{\textbf{Scores}} 
% & \textbf{Latency(ms)} & \multicolumn{3}{c}{\textbf{Scores}} \\
% \midrule
% SkyworkRM-Qwen3-8B & -- & .90 & .92 & .14 & 0.89 & -- & -- & -- \\
% \rowcolor{blue!6}
% Gnosis             & -- & \textbf{.95} & \textbf{.95} & \textbf{.09} & -- & -- & -- & -- \\
% \bottomrule
% \end{tabular}
% \end{table}

\begin{table}[t]
\centering
\small
\setlength{\tabcolsep}{1pt}
\renewcommand{\arraystretch}{0.9}
\caption{Correctness detection on Math-reasoning for \textbf{Qwen3 1.7B-Hybrid} backbone under two max response lengths (12k, 24k).
We compare \textbf{Gnosis} with \textbf{SkyworkRM-Qwen3-8B}, highlighting Gnosis’s near-constant latency and large speedups as response length increases.}
\label{tab:table4_speed}

\begin{tabular}{l|c|c c}
\toprule
\textbf{Method} & \textbf{Max len} &
\textbf{ Latency(ms) $\downarrow$} & \textbf{ AUROC $\uparrow$} \\
\midrule
SkyworkRM-Qwen3-8B & 12k & 930  & .90 \\
SkyworkRM-Qwen3-8B & 24k & 2465 & .88 \\

\arrayrulecolor{gray!60}\specialrule{0.15pt}{0pt}{1pt}\arrayrulecolor{black}

\rowcolor{blue!6}
Gnosis & 12k & \textbf{25}\,{\scriptsize($\times 37$)} & .95 \\
\rowcolor{blue!6}
Gnosis & 24k & \textbf{25}\,{\scriptsize($\times 99$)} & .94 \\
\bottomrule
\end{tabular}
\end{table}

\section{The Gnosis Mechanism}
\label{sec:gnosis}

We introduce \emph{Gnosis}, a lightweight self-awareness mechanism designed to retrofit frozen LLMs with introspection capabilities. Gnosis operates on the intuition that a model’s internal traces, its evolving hidden states and attention routing patterns, carry distinctive ``fingerprints'' of hallucination and reasoning errors. Unlike external judges that require separate, expensive decoding passes, Gnosis is a passive observer: it compresses the backbone's internal signals into compact descriptors and fuses them to predict a scalar correctness score. The architecture is explicitly designed so that its inference cost is independent of the sequence length, adding negligible overhead even for very long contexts.

\subsection{Problem Setup and Length-Invariant Inputs}
\label{sec:gnosis-setup}

% Let $x$ be an input prompt and $\hat{y} = (\hat{y}_1, \dots, \hat{y}_T)$ the answer generated autoregressively by a frozen backbone LLM, yielding a sequence of length $S$ tokens (prompt + answer). 

Let $x$ denote an input prompt of length $S_x$ and $\hat{y}$ the generated response of length $S_y$. The input to Gnosis is the concatenated sequence with a total size of $S = S_x + S_y$ tokens." The backbone has hidden dimension $D$, $L$ decoder layers, and $H$ attention heads per layer. During generation, we read only the final-layer hidden states $H^{\text{last}} \in \mathbb{R}^{S \times D}$ and the attention maps $\mathcal{A} = \{A_{\ell,h}\}_{\ell=1..L,\,h=1..H}$, where each $A_{\ell,h} \in \mathbb{R}^{S \times S}$ is the attention map of head $h$ in layer $\ell$.

Gnosis learns a verification function:
\begin{equation}
    \hat{p} = f_\phi\!\big(H^{\text{last}}, \mathcal{A}\big) \in [0,1],
\end{equation}
where $\hat{p}$ is the estimated probability that the generated answer is correct and $\phi$ are the parameters of Gnosis. The backbone LLM remains frozen throughout.

\paragraph{Fixed-Budget Compression.}
To decouple computational cost from sequence length $S$, we use a projection operator $\Pi$ that maps variable-length traces into fixed-size tensors:

\begin{itemize}[itemsep=0pt,parsep=0pt,topsep=0pt]
    \item \textbf{Hidden States.} The sequence $H^{\text{last}} \in \mathbb{R}^{S \times D}$ is interpolated and adaptively pooled along the sequence dimension to a fixed budget $K_{\text{hid}}$ (e.g.,$192$), yielding
    \begin{equation}
        \tilde{H} = \Pi_{\text{hid}}(H^{\text{last}}) \in \mathbb{R}^{K_{\text{hid}} \times D}.
    \end{equation}
    \item \textbf{Attention Maps.} Each attention map $A_{\ell,h} \in \mathbb{R}^{S \times S}$ is downsampled via adaptive pooling to a fixed grid size $k \times k$ (e.g., $k = 256$), giving a standardized set
    \begin{equation}
        \tilde{\mathcal{A}} = \{\tilde{A}_{\ell,h}\}_{\ell,h}, \quad \tilde{A}_{\ell,h} \in \mathbb{R}^{k \times k}.
    \end{equation}
\end{itemize}

All downstream encoders operate only on $\tilde{H}$ and $\tilde{\mathcal{A}}$ with fixed dimensions $(K_{\text{hid}}, D)$ and $(L, H, k, k)$, so the computational cost of Gnosis does not grow with $S$ and is negligible compared to the backbone; see Appendix~\ref{app:arch-overview} for architectural details.

\subsection{Hidden-State Circuit Encoder}
\label{sec:gnosis-hidden}

Standard confidence methods often rely on token probabilities (logits), which are poorly calibrated and only weakly aligned with correctness~\cite{ghasemabadi2025guided}. Gnosis instead learns from the backbone's internal representation, extracting correctness cues directly from the final-layer latent representations. A small encoder $\rho_{\text{hid}}$ maps this latent trace into a compact descriptor:
\begin{equation}
    z_{\text{hid}} = \rho_{\text{hid}}(\tilde{H}) \in \mathbb{R}^{D_{\text{HID}}}.
\end{equation}

\paragraph{Local Temporal Encoder.}
We treat $\tilde{H}$ as a temporal signal and apply a lightweight multi-scale 1D depthwise convolution over the sequence dimension to capture local dependencies and irregularities in the hidden trajectory.
% We treat $\tilde{H}$ as a temporal signal and apply a small multi-scale 1D convolutional block over the sequence dimension to capture short-range dependencies and local irregularities in the hidden trajectory. This block is lightweight and implemented with depthwise convolutions.

\paragraph{Global Set Encoder.}
To summarize the sequence into a compressed representation, we then apply a Set Transformer–style encoder~\citep{lee2019set}: Set Attention Blocks (SAB) followed by a Pooling-by-Multihead-Attention (PMA) block. This enables global interaction across all positions and aggregates the sequence into a small set of summary tokens, which are flattened and linearly projected to form the final hidden descriptor $z_{\text{hid}}$. Figure~\ref{fig:arch_detail} illustrates the detailed architecture design of Hidden Circuit Encoder. Appendix~\ref{app:hiddenencoder} details the encoder architecture, while Appendix~\ref{app:ablation-hidden} presents full design ablations.

\subsection{Attention Circuit Encoder}
\label{sec:gnosis-attn}

The attention stream $\tilde{\mathcal{A}}$ reveals layer- and head-level routing patterns that can indicate brittle reasoning or unstable focus, complementing the hidden-state descriptor. Rather than feeding raw attention weights into a large network, Gnosis summarizes each downsampled attention map $\tilde{A}_{\ell,h} \in \mathbb{R}^{k \times k}$ into a compact feature vector:
\begin{equation}
    \mathbf{v}_{\ell,h} = \Phi(\tilde{A}_{\ell,h}) \in \mathbb{R}^{d_{\text{grid}}}.
\end{equation}
Here $\Phi$ denotes our per-map feature extractor, which outputs a $d_{\text{grid}}$-dimensional summary for each attention map.

% \paragraph{Per-map Feature Extraction.}
% We implement $\Phi$ using two complementary approaches: (i) a lightweight CNN that learns local and global patterns from each attention map treated as an image, and (ii) an interpretable statistics-based extractor that computes predefined structural descriptors. We analyze each variant and their hybrid in Appendix~\ref{app:ablation-attn} (Table~\ref{tab:abl-attn-phi-all}). While the two extractors are competitive individually, their hybrid is most consistent across benchmarks; we therefore use $\left[\Phi_{\text{cnn}};\Phi_{\text{stat}}\right]$ in the final design.

% \paragraph{Interpretable Attention Statistics.}
% The statistics branch captures a small set of complementary signals: spectral entropy to distinguish sharp, confident heads from diffuse, noisy ones; radial band energies to summarize coarse vs.\ fine-grained attention texture; spatial centroids and variances to quantify where attention concentrates and how broadly it spreads; diagonal (and near-diagonal) mass to measure locality versus longer-range routing. Details of the statistics branch is provided in Appendix~\ref{app:attnencoder}.

\paragraph{Per-map Feature Extraction.}
We implement $\Phi$ using two complementary approaches: (i) a lightweight CNN that treats each attention map as an image and learns local-to-global patterns, and (ii) an interpretable statistics-based extractor that summarizes how attention is distributed, where it concentrates, and how local or long-range it is. Concretely, the statistics include simple measures of spread and texture (e.g., entropy- and frequency-based features), diagonal and near-diagonal mass to capture locality, and lightweight center-and-spread measures that describe the average location of attention and how widely it is dispersed. 

We ablate each variant and their hybrid in Appendix~\ref{app:ablation-attn} (Table~\ref{tab:abl-attn-phi-all}). While the two extractors are individually competitive, the hybrid is the most consistent across benchmarks; we therefore adopt $\left[\Phi_{\text{cnn}};\Phi_{\text{stat}}\right]$ in the final design. Full definitions of the statistics are provided in Appendix~\ref{app:attnencoder}.

% \paragraph{Circuit Grid and Encoder.}
% We stack the per-head feature vectors into an $L \times H$ ``circuit grid'':
% \begin{equation}
%     G \in \mathbb{R}^{L \times H \times d_{\text{grid}}}, \quad
%     G_{\ell,h,:} = \mathbf{v}_{\ell,h}.
% \end{equation}
% We add learned layer and head embeddings to encode depth and head index, and treat this grid as token. A shallow attention encoder $\rho_{\text{attn}}$, implemented as a few axial convolutional layers and a Pooling-by-Multihead-Attention block, processes $G$ to model cross-layer and cross-head dependencies, and collapses it into a single structural descriptor:
% \begin{equation}
%     z_{\text{attn}} = \rho_{\text{attn}}(G) \in \mathbb{R}^{D_{\text{ATT}}}.
% \end{equation}
% All computation operates on fixed dimensions $(L, H, d_{\text{grid}})$ and is independent of $S$. Detailed architecture choices and ablations are defered to Appendix~\ref{app:ablation-attn}.

% \paragraph{Encoding Cross-Layer and Cross-Head Structure.}

\begin{figure*}[!t]
    \centering
    \includegraphics[width=1\textwidth]{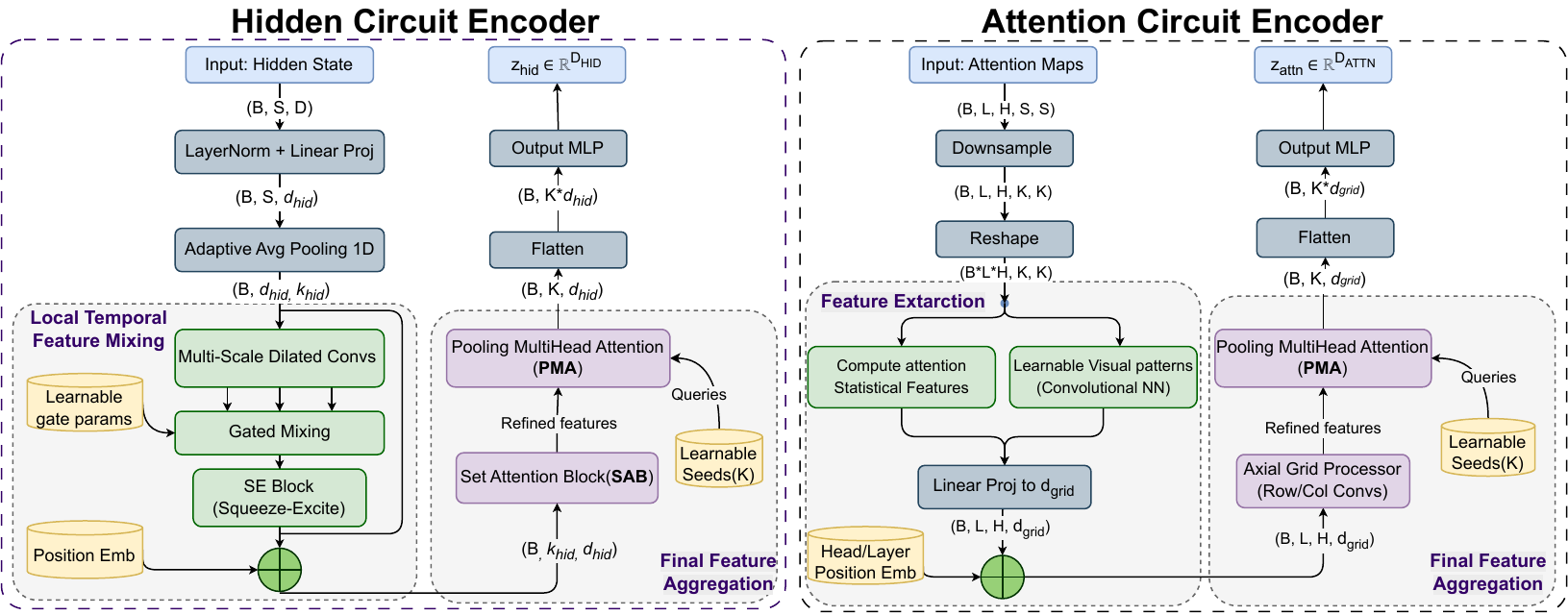}
    \vspace{-1em}
    \caption{\textbf{Gnosis Encoder Architecture Details.}
    \textbf{Hidden Circuit (left):} project and adaptively pool the hidden-state trace, apply multi-scale dilated temporal mixing, then use lightweight \emph{attention-based pooling} (SAB$\rightarrow$PMA) to produce a compact descriptor $z_{\text{hid}}$.
    \textbf{Attention Circuit (right):} downsample each layer--head attention map to a fixed $k{\times}k$ grid, extract per-map CNN+statistics features, mix across the layer$\times$head grid with a lightweight axial processor, and pool (PMA) to obtain $z_{\text{attn}}$. Appendix~\ref{app:arch-overview} gives a detailed description of the encoder design, and Appendix~\ref{app:all_ablations} includes the complete set of architecture and design ablations.}
    \label{fig:arch_detail}
\end{figure*}

\paragraph{Cross-Head and Cross-Layer Encoding.}
We arrange the per-head summaries into an $L \times H$ layer--head grid:
\begin{equation}
    G \in \mathbb{R}^{L \times H \times d_{\text{grid}}}, \quad
    G_{\ell,h,:} = \mathbf{v}_{\ell,h}.
\end{equation}
We add learned layer and head embeddings to preserve depth and head identity. We then treat the $L \times H$ entries as grid tokens. A lightweight encoder $\rho_{\text{attn}}$ mixes information across layers and heads using a few axial convolutional layers. This design is substantially cheaper than full global self-attention over the grid. Finally, we apply Pooling-by-Multihead-Attention (PMA) to aggregate the grid into a single descriptor:
\begin{equation}
    z_{\text{attn}} = \rho_{\text{attn}}(G) \in \mathbb{R}^{D_{\text{ATT}}}.
\end{equation}
Because this stage operates on fixed dimensions $(L, H, d_{\text{grid}})$, the Gnosis-side compute is independent of the original sequence length $S$. Figure~\ref{fig:arch_detail} illustrates the detailed architecture design of Attention Circuit Encoder. Detailed architecture choices and ablations are deferred to Appendix~\ref{app:ablation-attn}.

\subsection{Gated Fusion and Correctness Prediction}
\label{sec:gnosis-fusion}

Gnosis fuses the hidden and attention descriptors into a single vector and maps it to a correctness probability. We concatenate the two descriptors
\begin{equation}
    z = [z_{\text{hid}}; z_{\text{attn}}],
\end{equation}
and feed the result into a small gated MLP head. The final correctness estimate is
\begin{equation}
    \hat{p} = \sigma\!\big(\mathrm{GatedMLP}_{\phi}(z)\big),
\end{equation}
where $\mathrm{GatedMLP}_{\phi}$ is a lightweight gated MLP and $\sigma$ is the sigmoid. This head lets Gnosis adaptively weight hidden versus attention features on a per-example basis(e.g., leaning more on attention for reasoning traces and more on hidden states for factual recall). The architecture is intentionally small: Gnosis adds only $\sim$5M parameters, making it $\sim$1000$\times$ smaller than 8B reward models and dramatically smaller than Gemini 2.5 pro as judge.

\subsection{Training}
\label{sec:gnosis-training}
A key advantage of Gnosis is that it can be trained without costly-annotated data. For each backbone, we generate answers on the training sets and label correctness by comparing predictions to ground-truth answers. This yields a binary classification dataset:
\[
    \mathcal{D} = \{(H^{\text{last}}_i, \mathcal{A}_i, y_i)\}_{i=1}^N,
\]
where $y_i \in \{0,1\}$ indicates whether the verifier judged the $i$-th generation as correct. Gnosis is trained to minimize binary cross-entropy:
\[
    \mathcal{L}(\phi)
    = -\mathbb{E}_{(H^{\text{last}}, \mathcal{A}, y) \sim \mathcal{D}}
    \big[\, y \log \hat{p} + (1-y)\log(1-\hat{p}) \,\big],
\]
with $\hat{p} = f_{\phi}(H^{\text{last}}, \mathcal{A})$. The backbone is frozen; gradients flow only through the Gnosis encoders and fusion head. 

% \paragraph{Complexity and efficiency.}
% Because of the length-to-budget projection ($T_{\text{fix}}$, $k$) and fixed circuit grid size $(L, H)$, the forward pass of Gnosis has complexity
% \[
%     O\big(T_{\text{fix}} D\big) + O\big(L H d_{\text{stat}}\big),
% \]
% which is \emph{constant} with respect to the sequence length $S$. In practice, Gnosis adds negligible end-to-end latency (on the order of a few milliseconds per example) and uses only $\sim$5M parameters, making it orders of magnitude smaller than typical 8B-parameter reward models while avoiding any additional autoregressive decoding.

\begin{figure*}[t]
    \centering
    \includegraphics[width=\textwidth]{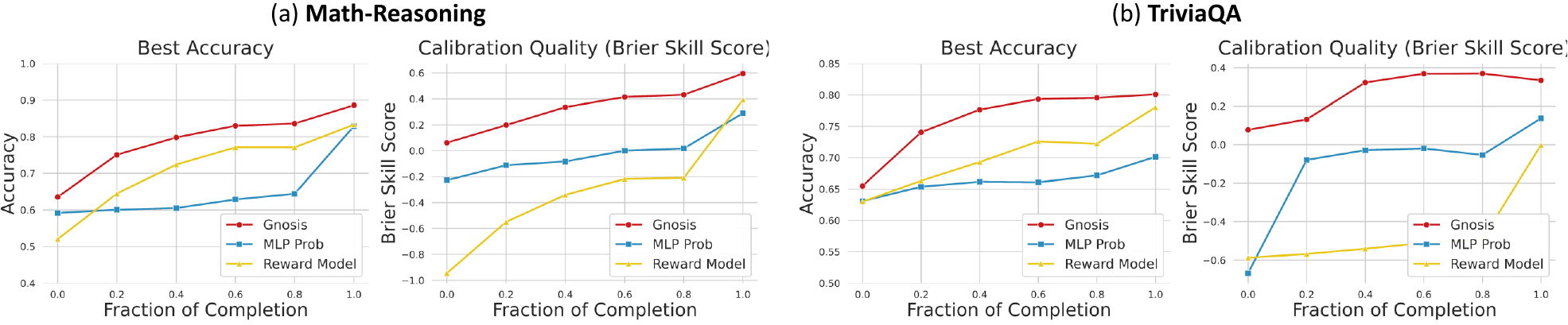}
    \vspace{-1.8em}  % tighten space between image and caption
    \caption{
    \textbf{Early Correctness Prediction on Math-Reasoning.} Gnosis (red) achieves higher accuracy 
    and better calibration than both MLP-Prob (blue) and a reward model \textsc{SkyworkRM-Qwen3-8B} (yellow). 
    Notably, after seeing 40\% of the completion, Gnosis already matches the full-solution 
    performance of the other methods.
    }
    \label{fig:early_pred}
\end{figure*}

\begin{figure*}[t]
    \centering
    \includegraphics[width=\textwidth]{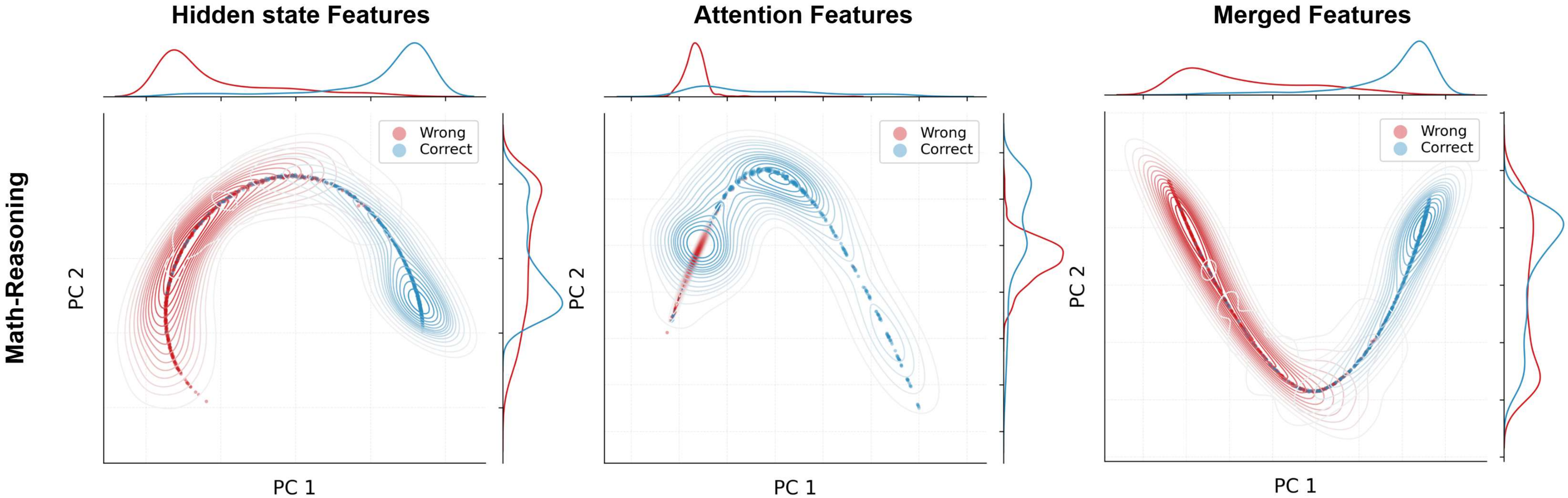}
    \vspace{-1.8em}  % tighten space between image and caption
    \caption{
    \textbf{2D Embeddings of Features Learned by Gnosis on Math-Reasoning.} We show dimensionality-reduced embeddings of hidden-state features (left), attention features (middle), and their merged features (right), with KDE contours and marginal densities for wrong (red) and correct (blue) answers. Hidden features exhibit the clearest separation, attention features show a weaker but still clear separation, and the merged space yields the sharpest overall discrimination between correct and wrong solutions.
    }
    \label{fig:disturbution_plot_math}
\end{figure*}

% \begin{figure*}[t]
%     \centering
%     \includegraphics[width=\textwidth]{figures/Gnosis_probdisturb.pdf}
%     \vspace{-2em}  % tighten space between image and caption
%     \caption{
%     \textbf{Predicted Correctness Score Distributions.} Gnosis (top) displays sharp, bimodal separation between correct (blue) and wrong (red) answers. In contrast, the larger Skywork model (bottom) exhibits diffuse distributions with significant overlap, reflecting higher uncertainty.
%     }
%     \label{fig:prob_disturb}
% \end{figure*}

\section{Experiments and Results}

We evaluate \textbf{Gnosis} in three practical regimes: 
(i) \emph{self-judgment}, where each Gnosis head scores generations from its own frozen backbone; 
(ii) \emph{sibling-model judgment}, where a small head serves as a lightweight reward model for larger family members; 
and (iii) \emph{early correctness prediction}, where Gnosis is queried on partial completions to support compute-aware control.

\subsection{Experimental Setup}
\textbf{Backbones.}
We apply Gnosis to five frozen LLMs: Qwen3 1.7B-Hybrid, Qwen3 4B-Thinking, Qwen3 4B-Instruct, Qwen3 8B-hybrid\cite{yang2025qwen3} and OpenAI gpt-oss-20B\cite{agarwal2025gpt}. The backbone weights and decoding settings are never updated.

\textbf{Training Data.} We train one correctness head per backbone on a mixed math–trivia corpus to cover both multi-step reasoning and open-domain factual recall.  
For math, we use the English portion of DAPO-Math-17k ($\sim$14k competition-style problems with numeric or symbolic answers\cite{yu2025dapoopensourcellmreinforcement}).  
For QA, we subsample 40k questions from a 118k-item TriviaQA training set\cite{2017arXivtriviaqa} to retain broad coverage while keeping training compact. We generate two completions per math prompt to capture diverse reasoning trajectories and increase correct/incorrect label variety under the same question, and one completion per trivia prompt since answers are shorter and often less ambiguous.  
We extract final answer, label correctness by comparing to the ground-truth, and discard outputs without valid answers.  
This yields a balanced, fully automated training set that requires no human annotation.

\textbf{Training Details and Cost.}
We train each head for two epochs over this mixed dataset using Adam with a learning rate of $1\times 10^{-4}$. Because the backbone is frozen and all feature extractors operate at a fixed budget independent of sequence length, training is lightweight.
For the largest backbone (gpt-oss-20B MoE), the full pipeline, data generation and training finishes in roughly 12 hours on $2\times$A100 80\,GB GPUs, corresponding to \$25 in cloud cost.
Smaller backbones train faster.

\textbf{Benchmarks.}
For each benchmark, we prompt each frozen backbone to generate a solution with a maximum budget of \textbf{12k tokens}, and retain only question--answer pairs with a valid final answer for evaluation.
We evaluate Gnosis on three disjoint domains: \textbf{Math-Reasoning} (AMC12 2022/2023\cite{aimo2024validationamc}, AIME 2024/2025\cite{aime24,aime25}, HMMT Feb 2025\cite{balunovic_srimatharena_2025}), \textbf{Open-Domain QA} (18k held-out TriviaQA questions with no overlap with training), and \textbf{Academic Knowledge Reasoning} (MMLU-Pro\cite{wang2024mmlu}). Together, these benchmarks stress multi-step reasoning, hallucination detection on short factoid answers, and out-of-distribution generalization. We report detailed backbone-level outcome statistics (accuracy, hallucination, and non-response rates) in Appendix~\ref{app:backbone-outcomes}. Additional benchmark details are provided in Appendix~\ref{app:additional-exp-setup}.

\textbf{Metrics.}
We treat correctness prediction as binary classification and report \textbf{AUROC} and \textbf{AUPR} under two complementary labelings (\textbf{AUPR-c}: \emph{correct} as positive; \textbf{AUPR-e}: \emph{incorrect} as positive), together with calibration metrics \textbf{Brier Skill Score (BSS)} and \textbf{Expected Calibration Error (ECE)}. AUROC/AUPR measure discriminative ranking under class imbalance, whereas BSS/ECE assess the quality and calibration of predicted probabilities. See Appendix~\ref{app:additional-exp-setup} for extended interpretations.

\textbf{Baselines.}
We compare against four baseline families. \textbf{(1) Statistical internal scores} are training-free indicators computed from the backbone’s own outputs, reported in Table~\ref{table1} as \emph{Logit Entropy}, \emph{Mean Token Prob}, and \emph{Attn Eigenvalue Score}\cite{sriramanan2024llmcheck}. \textbf{(2) Trajectory/spectral internal indicators} summarize cross-layer hidden-state dynamics, reported as \emph{CoE--R} and \emph{CoE--C}\cite{wang2025latentcoe}. \textbf{(3) External judges} include two open-source reward models that are state-of-the-art on public reward-model benchmarks\cite{malik2025rewardbench}, \emph{SkyworkRM-Llama3.1-8B} and the family-aligned \emph{SkyworkRM-Qwen3-8B}\cite{liu2025skywork}, as well as \emph{Gemini 2.5 Pro} used as a judge (the Gemini judging prompt is provided in the Appendix~\ref{app:gemini}); all are reported in Table~\ref{table1}. \textbf{(4) A Learnable probe}\cite{zhang2025reasoning} that observes only the final answer token’s hidden state is reported separately on Qwen3 1.7B in Table~\ref{table2} to isolate the limitations of single-token probing.

\subsection{Self-Judgment}
\label{sec:self-judgment}
We evaluate Gnosis in the standard \emph{self-judgment} setting: for each backbone, the model generates answers to the benchmark questions, and the verification method predicts the correctness of these specific generations. As shown in Tables~\ref{table1} and \ref{table2}, Gnosis consistently outperforms training-free baselines and large external judges across all tested domains.

\textbf{Superiority Over Internal Baselines and Probes.}
Across Math Reasoning and Open-Domain QA, Gnosis effectively solves the miscalibration of standard confidence metrics. It consistently lifts AUROC from the mid-0.7s (typical of Logit Entropy) to \textbf{0.95--0.96} while roughly doubling the BSS, turning negative calibration scores into strongly positive ones. Crucially, Gnosis outperforms the learned \emph{MLP-Prob} final-token probe by \textbf{7--18 AUROC points} across benchmarks. This consistent gap confirms that correctness is a property of the full generation trajectory, specifically the distributed hidden-state dynamics and attention patterns, rather than a state localized to the final token.

\textbf{Efficiency vs.\ Scale:}
With only $\sim$5M parameters and negligible overhead from its fixed-budget projection, 
Gnosis matches or exceeds state-of-the-art \textbf{Skywork 8B Reward Models} ($\sim$1000$\times$ larger) and the proprietary \textbf{Gemini 2.5 Pro}. 
This is notable because Gnosis adds no independent world knowledge. 
Rather than fact-checking with massive parametric memory, it detects the signatures of hallucination and reasoning error in the backbone’s internal traces. 
On complex Math Reasoning, Gnosis surpasses both large judges. 
It also outperforms Gemini on \textbf{MMLU-Pro}, a domain it was not explicitly trained on, suggesting that it learns transferable error patterns instead of task-specific cues. 
Additionally, Gnosis maintains near-constant $\sim$25ms latency and achieves roughly \(37\times\) and \(99\times\) speedups over the 8B reward model when judging answers of length 12k and 24k tokens, respectively (Table~\ref{tab:table4_speed}).
These results show that intrinsic self-verification can be both more scalable and far cheaper than external oversight.

% Figure~\ref{fig:prob_disturb} visualizes this advantage, showing that Gnosis produces sharp, \emph{bimodal distributions} with clear peaks at 0 (incorrect) and 1 (correct). In stark contrast, the Skywork 8B Reward Model yields diffuse distributions often clustered around 0.5--0.6, reflecting deep uncertainty. This explains why Gnosis achieves superior calibration (BSS): it successfully pushes predictions to the correct extremes where larger models struggle to commit.

Figure~\ref{fig:prob_disturb} compares the predicted correctness score distributions of Gnosis and the Skywork 8B Reward Model. 
Gnosis shows sharp, \emph{bimodal} peaks near 0 (incorrect) and 1 (correct), 
whereas Skywork produces broader, overlapping scores that often cluster around 0.5--0.6. This separation aligns with Gnosis’s stronger calibration (BSS) and its tendency to assign more decisive probabilities.

\vspace{-0.8em}
\subsection{Cross-Scale: Zero-Shot Reward Modeling}
\vspace{-0.5em}
\label{sec:cross-model}

We introduce ``Sibling Modeling'', where we train Gnosis on a small, cheap backbone and deploy it to judge larger family members without fine-tuning. Table~\ref{tab:sibling-judging} highlights a striking outcome: a head trained on a 1.7B backbone transfers effectively to 4B and 8B siblings across all evaluated domains. On Math Reasoning, for instance, it achieves 0.93 AUROC, nearly matching the 0.96 achieved by a self-trained head. Notably, this transferred 1.7B head still consistently outperforms the Skywork 8B Reward Model across all tested backbones, proving that our tiny zero-shot verifier is more reliable than a massive external judge. This broad transferability implies that hallucination manifests as a structural invariant across model scales, offering a ``free lunch'' where a single small head serves as a supervisor for an entire model family. We observe that this transfer is most effective when models share a similar generation style; while Gnosis robustly handles differences in size, it performs best when the models also align in their formatting (e.g., transferring between thinking models rather than thinking-to-Instruct).
\vspace{-0.4em}
\subsection{Early Error Detection}
\label{sec:early-prediction}
\vspace{-0.5em}
Because Gnosis processes internal traces into fixed-size descriptors, it can evaluate partial generations natively. \emph{Crucially, this capability is emergent:} although Gnosis is trained exclusively on complete trajectories, it generalizes zero-shot to partial prefixes without any additional fine-tuning. Figure~\ref{fig:early_pred} illustrates that on both Math Reasoning and TriviaQA, Gnosis reaches near-peak accuracy and positive BSS after observing only \textbf{40\%} of the generation. In contrast, external reward models and single-token learnable probe typically require the full response to stabilize. This enables aggressive \textbf{compute-aware control policies}: generated chains-of-thought can be terminated immediately if the internal ``hallucination alarm'' triggers, preventing wasted compute on failing paths, or the system can automatically escalate the query to a stronger model upon detecting that the current backbone is incapable of answering correctly.

% ===================== Main paper =====================
% \section{Ablations and Analysis}
% \label{sec:ablations}

% To validate the architectural design of Gnosis, we conducted comprehensive ablation studies analyzing the individual contributions of the hidden and attention streams, as well as the specific design choices within their respective encoders. We investigated the impact of attention feature representations (spectral statistics vs. learned CNNs), grid topology, and aggregation strategies. Similarly, for the hidden-state circuit, we analyzed the efficiency and performance trade-offs of local temporal processing and global set-based aggregation.

% Our results confirm that fusing both hidden and attention signals yields the highest reliability ($0.97$ AUROC), significantly outperforming single-stream variants2. Furthermore, we found that explicitly computed spectral statistics for attention maps outperform learned visual features, and that a hybrid local-global processing stack boosts the hidden state encoder's performance while remaining computationally efficient33. Detailed experimental results, architectural diagrams, and ablation tables are provided in Appendix A.
\vspace{-0.5em}
\section{Ablations and Analysis}
\vspace{-0.5em}
\label{sec:ablations}

We highlight the key ablation insights that motivate Gnosis design, and defer comprehensive studies to Appendix~A.

\textbf{Hidden vs.\ Attention Circuits.}
% Gnosis combines a hidden-state circuit and an attention circuit. Appendix~Table~\ref{tab:abl-hidden-attn-all} shows that each stream is useful but their strengths differ by domain. On TriviaQA, the hidden-only model clearly outperforms attention-only, and the fused model closely tracks the hidden stream, indicating that hidden states carry most of the reliability signal for short factual recall. On Math Reasoning, both single-stream variants are strong, while fusing them yields a noticeable gain, suggesting complementary cues. On MMLUPro, attention-only is competitive with (and slightly stronger than) hidden-only, and fusion preserves the best overall performance. 
Gnosis fuses a hidden-state and an attention circuit. Appendix~Table~\ref{tab:abl-hidden-attn-all} shows that both streams help, but attention is most useful for long-form reasoning: on TriviaQA, hidden-only dominates and fusion adds little, indicating hidden states carry most short factual reliability; on Math Reasoning and MMLU-Pro, attention-only is strong (even slightly better than hidden-only on MMLU-Pro), and fusion yields the best overall performance, suggesting complementary structural cues that emerge in longer reasoning.

We visualize this behavior on Math-Reasoning in Figure~\ref{fig:disturbution_plot_math}; analogous feature-distribution plots for the other domains are provided in the appendix Figure~\ref{fig:scatterplot_all}. Taken together, these results support that hidden states provide a broad, robust signal across domains, while attention routing contributes more strongly on reasoning-heavy tasks and less on short factual QA; combining both is the most reliable overall.

\textbf{Attention Map Extractor.}
We ablate how to summarize each downsampled attention map with a lightweight CNN, predefined fixed statistics, and their hybrid. Appendix~Table~\ref{tab:abl-attn-phi-all} shows that the three variants perform similarly on Math, while on TriviaQA the CNN-based variants are stronger than statistics alone, and on MMLU-Pro the CNN+Stats hybrid is the most consistent. Based on these, we adopt the \textbf{CNN+Stats} design in the final model.

% \paragraph{Additional Ablations.}
% We further validate the design path behind Gnosis with targeted ablations across both streams. Appendix~A details how choices in the attention circuit (grid mixing, identity embeddings, pooling strategy, layer selection, and map downsampling) affect performance (Appendix~Tables~\ref{tab:attn_ablation_components} and \ref{tab:attn_ablation_hyperparams}), alongside complementary studies of the hidden circuit’s local--global encoder and sizing trade-offs (Appendix~Tables~\ref{tab:hidden-ablation-components} and \ref{tab:hidden-ablation-sizing}).

\textbf{Additional Ablations.}
We further validate the design path behind Gnosis with targeted ablations across both streams. On the attention stream, Appendix~\ref{app:ablation-attn} examines how grid mixing, identity embeddings, pooling strategy, layer selection, and map downsampling affect performance (Appendix~Tables~\ref{tab:attn_ablation_components} and \ref{tab:attn_ablation_hyperparams}). On the hidden stream, we isolate architectural value by including a simple pooled-MLP baseline that naively pools final-layer hidden states before an MLP (Row G in Table~\ref{tab:hidden-ablation-components}), alongside broader studies of the local--global encoder design and sizing trade-offs (Appendix~Tables~\ref{tab:hidden-ablation-components} and \ref{tab:hidden-ablation-sizing}).
\vspace{-0.5em}
\section{Discussion and Limitations}
\vspace{-0.5em}
Gnosis provides a highly efficient framework for self-evaluation and ``sibling modeling,'' where a small head trained on a compact model effectively judges larger models within the same family. This architecture further supports compute-aware control by enabling early error detection on partial generation traces. However, a key limitation is that Gnosis is designed as a self-awareness mechanism rather than a general-purpose reward model; while it transfers robustly to siblings, it is not capable of acting as a universal zero-shot judge for every model, particularly those with unrelated architectures or differing generation styles (e.g., transferring between Thinking and Instruct models).
\vspace{-0.5em}
\section{Conclusion}
\vspace{-0.5em}
We introduced Gnosis, a lightweight mechanism that allows frozen LLMs to detect their own errors by interpreting internal hidden and attention traces rather than relying on external judges. Despite adding only $\sim$5M parameters, Gnosis consistently outperforms billion-parameter reward models and Large proprietary models like Gemini 2.5 Pro, demonstrating that high-fidelity correctness signals are intrinsic to the generation process. This approach establishes a new standard for compute-efficient reliability, enabling self-verifying systems that can detect failing trajectories with negligible overhead.

\bibliography{custom}

\newpage
\appendix
\onecolumn
\section{Architecture Overview}
\label{app:arch-overview}

\begin{figure*}[t]
    \centering
    \includegraphics[width=\textwidth]{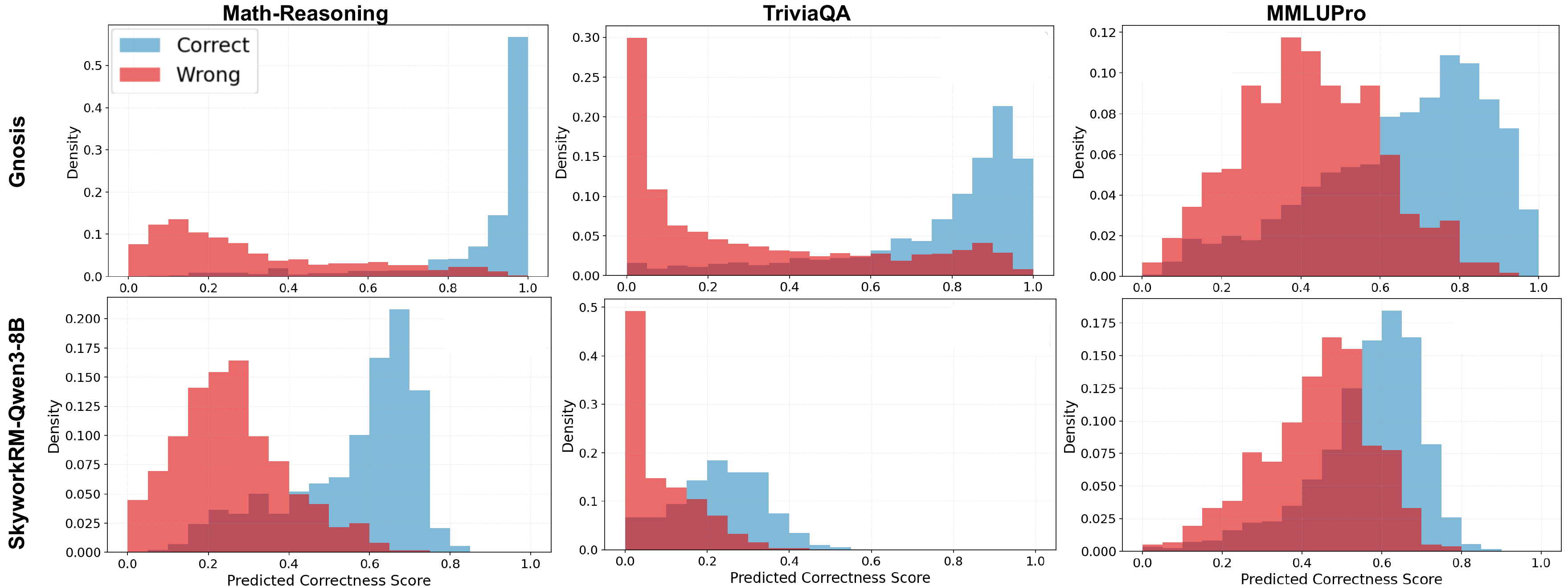}
    \vspace{-2em}  % tighten space between image and caption
    \caption{
    \textbf{Predicted Correctness Score Distributions.} Gnosis (top) displays sharp, bimodal separation between correct (blue) and wrong (red) answers. In contrast, the larger Skywork model (bottom) exhibits diffuse distributions with significant overlap, reflecting higher uncertainty.
    }
    \label{fig:prob_disturb}
\end{figure*}

\begin{table*}[t]
\centering
\footnotesize
\setlength{\tabcolsep}{2pt}
\renewcommand{\arraystretch}{1}
\caption{\textbf{Impact of Dual-Stream Architecture across Benchmarks.}
Comparison of single-stream variants against the Full Gnosis model.
While hidden states provide a strong signal, fusing them with the attention circuit consistently yields the best performance.}
\label{tab:abl-hidden-attn-all}
\begin{tabular}{ll|ccc|ccc|ccc}
\toprule
& & \multicolumn{3}{c|}{\textbf{Math Reasoning}} & \multicolumn{3}{c|}{\textbf{TriviaQA}} & \multicolumn{3}{c}{\textbf{MMLU-Pro}} \\
\textbf{Model Variant} & \textbf{Input Signals} & \textbf{AUROC} & \textbf{AUPR-c} & \textbf{AUPR-e} & \textbf{AUROC} & \textbf{AUPR-c} & \textbf{AUPR-e} & \textbf{AUROC} & \textbf{AUPR-c} & \textbf{AUPR-e} \\
\midrule
\rowcolor{blue!10} \textbf{\textsc{Both (Gnosis)}} & \textbf{Fused} ($[z_{\text{hid}}; z_{\text{attn}}]$) & \textbf{0.95} & \textbf{0.95} & \textbf{0.94} & \textbf{0.87} & \textbf{0.79} & \textbf{0.92} & \textbf{0.80} & \textbf{0.90} & \textbf{0.56} \\
\midrule
\textsc{Attention-only} & Attention Maps ($z_{\text{attn}}$) & 0.92 & 0.93 & 0.90 & 0.78 & 0.66 & 0.86 & 0.80 & 0.90 & 0.56 \\
\textsc{Hidden-only} & Hidden States ($z_{\text{hid}}$) & 0.92 & 0.92 & 0.91 & 0.87 & 0.77 & 0.92 & 0.78 & 0.89 & 0.53 \\
\bottomrule
\end{tabular}
\end{table*}

Figure~\ref{fig:arch_detail} illustrates the detailed internal components of the Gnosis Mechanism. The architecture consists of two parallel streams that process the frozen backbone's internal traces to extract reliability signals efficiently:

\subsection{Hidden Circuit Encoder.}
\label{app:hiddenencoder}
This stream processes the sequence of hidden states
$H_{\text{last}} \in \mathbb{R}^{S \times D}$. To avoid the added cost and memory of storing intermediate states for every token, 
we use only the final-layer hidden state. 
We show this choice remains strongly predictive.
To handle variable lengths while maintaining a fixed compute budget, the sequence is first projected and pooled into a fixed number of tokens.
It then passes through a \emph{Local Temporal Feature Mixing} stage (Phase~1) using multi-scale dilated convolutions and Squeeze-and-Excitation (SE) blocks to capture local dependencies and reweight informative channels.
Finally, a \emph{Global Set Encoder} (Phase~2) utilizes a Set Attention Block (SAB) followed by Pooling by Multihead Attention (PMA) to aggregate the sequence into a compact descriptor
$z_{\text{hid}} \in \mathbb{R}^{D_{\text{HID}}}$.

\subsection{Attention Circuit Encoder.}
\label{app:attnencoder}
This stream processes the collection of attention maps
$\{A_{\ell,h}\}_{\ell=1..L,\,h=1..H}$ from a frozen backbone.
To make computation invariant to the original context length, we downsample
each map to a fixed grid
$\tilde{A}_{\ell,h} \in \mathbb{R}^{k \times k}$.
We then summarize each downsampled map into a compact per-head descriptor
$\mathbf{v}_{\ell,h} = \Phi(\tilde{A}_{\ell,h}) \in \mathbb{R}^{d_{\text{grid}}}$.
These per-head descriptors are arranged as a layer--head grid and processed
by an Axial Grid Processor to model inter-layer and inter-head dependencies,
followed by PMA to obtain the final attention descriptor
$z_{\text{attn}} \in \mathbb{R}^{D_{\text{ATT}}}$.

% \paragraph{Per-Map Feature Extraction Variants.}
% As described in Section~\ref{sec:gnosis-attn}, we implement $\Phi$ using two
% alternatives:
% (i) a lightweight CNN that treats each attention map as an image and learns
% local-to-global patterns, and
% (ii) an interpretable statistics-based extractor that computes predefined
% structural descriptors.
% We compare these variants in Table~\ref{tab:abl-attn-phi-all} and observe similar performance.
% To keep Gnosis maximally interpretable we adopt the predefined statistics extractor in our final design.

\paragraph{Per-Map Feature Extraction Variants.}
As described in Section~\ref{sec:gnosis-attn}, we implement $\Phi$ using two
alternatives: (i) a lightweight CNN that treats each attention map as an image and learns
local-to-global patterns, and (ii) an interpretable statistics-based extractor that computes predefined
structural descriptors. We compare these variants and their hybrid in
Table~\ref{tab:abl-attn-phi-all}. While CNN-only and Stats-only are individually competitive,
the hybrid is the most consistent across domains. Unless otherwise stated, we therefore adopt
\textbf{CNN+Stats} as the default Gnosis configuration. The Stats-only variant remains a strong,
more interpretable alternative.

\paragraph{Interpretable Attention Statistics.}
The statistics branch summarizes how attention is distributed, how local or long-range it is, and where it tends to concentrate on the map, using a small set of predefined descriptors. Concretely, for each downsampled map $\tilde{A}_{\ell,h}$ we compute:

\begin{itemize}
    \item \textbf{Entropy.}
    We compute \emph{map entropy} together with \emph{row} and \emph{column entropies}.
    Map entropy reflects the overall dispersion of attention mass (focused vs.\ diffuse).
    Row and column entropies provide axis-specific views of this dispersion,
    indicating whether spread is driven primarily by query positions (rows)
    or key positions (columns).
    Together, these metrics offer a direct, interpretable summary of attentional
    diffusion and stability.

    \item \textbf{Spectral Texture.}
    We compute spectral entropy and the relative energy from the 2D Fourier spectrum of $\tilde{A}_{\ell,h}$.
    These features summarize whether attention exhibits coherent, structured
    patterns (low-frequency dominance) or becomes fragmented and noisy
    (elevated high-frequency energy and higher spectral entropy).

    \item \textbf{Locality via Diagonal Structure.}
    We measure diagonal and near-diagonal mass through the diagonal ratio and
    diagonal-band energies. This provides a simple proxy for locality versus
    longer-range routing, which often correlates with coherent step-wise reasoning behavior.

    \item \textbf{Center and Spread on the Map.}
    We compute lightweight center-and-width measures to describe the average
    location of attention mass and how widely it is dispersed across the grid.
    
\end{itemize}

The two descriptors are concatenated and passed through a lightweight gated MLP to produce a scalar correctness logit, which is converted to a probability via a sigmoid. Because both encoders operate on fixed-size summaries, Gnosis runs at effectively constant cost in sequence length and can be queried on partial chains of thought.

\section{Comprehensive Ablations and Analysis}
\label{app:all_ablations}
\subsection{Hidden vs.\ Attention Circuits}
\label{app:ablation-dual-stream}

To quantify the distinct contributions of the internal representations, we trained single-stream variants of Gnosis and compared them to the full dual-stream model on Qwen3 1.7B.

As reported in Table~\ref{tab:abl-hidden-attn-all}, the HIDDEN-ONLY model already provides a strong correctness signal across benchmarks. On Math Reasoning, both single-stream variants reach 0.92 AUROC, while fusing $z_{\text{hid}}$ and $z_{\text{attn}}$ improves performance to 0.95 AUROC. On TriviaQA, the hidden stream remains stronger than attention alone, and fusion achieves the best overall performance.
On MMLU-Pro, attention-only is slightly stronger than hidden-only, while fusion matches the best result. These results confirm that attention contributes complementary structural cues that help maximize performance when combined with hidden representations.

\subsection{Attention Circuit Encoder Ablations}
\label{app:ablation-attn}

To determine the optimal architecture for the attention circuit, we systematically
investigated three key design components: the input feature representation for each
attention map, the grid topology for mixing information across layers and heads,
and the final aggregation strategy.
Tables~\ref{tab:abl-attn-phi-all}, \ref{tab:attn_ablation_components}, and \ref{tab:attn_ablation_hyperparams} detail this investigation.
We select the configuration highlighted in \textbf{Row~A} of Tables~\ref{tab:attn_ablation_components} and \ref{tab:attn_ablation_hyperparams} as it achieves the best balance of
accuracy and parameter efficiency.

\textbf{Feature Input Representation.}
We first assessed how best to encode individual attention maps.
Table~\ref{tab:abl-attn-phi-all} compares a lightweight learned CNN, predefined statistics, and their combination across benchmarks.
The predefined statistics remain competitive with the CNN (e.g., identical 0.92 AUROC on Math Reasoning), while providing a more interpretable per-map representation.
Combining CNN and statistics yields comparable performance overall and modest gains on MMLU-Pro (0.80 AUROC).

\textbf{Grid Topology and Layer/Head Identity.}
We next evaluated how to process the collection of extracted map features.
As shown in Table~\ref{tab:attn_ablation_components}, removing the grid mixing
entirely (Row~B) reduces performance, indicating that
individual attention heads are not independent predictors; their interactions matter.
Replacing our lightweight \emph{Axial Convolutions} with a heavy Global Transformer
(Row~C) increases parameters four-fold without improving AUROC, validating the
efficiency of the axial design.
Furthermore, removing the learned layer and head embeddings (Row~D) degrades
performance, confirming that the model relies on knowing \emph{where}
a specific activation pattern occurred within the LLM's depth and breadth.

\textbf{Aggregation Strategy.}
Finally, we analyzed how to summarize the grid into a fixed vector.
Replacing our query-based \emph{Pooling by Multihead Attention (PMA)} with simple
mean pooling (Row~H in Table~\ref{tab:attn_ablation_components}) causes a sharp drop in accuracy.
This suggests that Gnosis benefits from learning specific "reliability prototypes"
(via PMA seeds) rather than uniformly averaging all attention circuits, likely
because only a small subset of heads carry high-fidelity correctness signals.

\begin{table*}[t]
\centering
\footnotesize
\setlength{\tabcolsep}{3pt}
\renewcommand{\arraystretch}{1.1}
\caption{\textbf{Impact of Per-Map Feature Extractor Choices across Benchmarks.}
Comparison of learnable CNN, predefined statistics, and their combination for the attention per-map extractor $\Phi$.
Across benchmarks, the CNN and statistics variants are competitive, while their hybrid is the most consistent overall; we adopt the CNN+Stats design in our final model.}
\label{tab:abl-attn-phi-all}
\begin{tabular}{l|ccc|ccc|ccc}
\toprule
& \multicolumn{3}{c|}{\textbf{Math Reasoning}} & \multicolumn{3}{c|}{\textbf{TriviaQA}} & \multicolumn{3}{c}{\textbf{MMLU-Pro}} \\
\textbf{Model Variant} &
\textbf{AUROC} & \textbf{AUPR-c} & \textbf{AUPR-e} &
\textbf{AUROC} & \textbf{AUPR-c} & \textbf{AUPR-e} &
\textbf{AUROC} & \textbf{AUPR-c} & \textbf{AUPR-e} \\
\midrule

\rowcolor{blue!10}
\textbf{\textsc{CNN+Stats (Final)}} &
0.92 & 0.93 & 0.90 &
0.78 & 0.65 & 0.86 &
0.80 & 0.90 & 0.56 \\
\midrule

\textsc{CNN-only} &
0.92 & 0.93 & 0.88 &
0.79 & 0.66 & 0.86 &
0.79 & 0.90 & 0.56 \\

\textsc{Stats-only} &
0.92 & 0.93 & 0.90 &
0.75 & 0.62 & 0.83 &
0.76 & 0.88 & 0.53 \\

\bottomrule
\end{tabular}
\end{table*}

\begin{table*}[t]
\centering
\small
\setlength{\tabcolsep}{4pt}
\renewcommand{\arraystretch}{1.1}

% LEFT TABLE: Components and Topology
\begin{minipage}[t]{0.58\linewidth}
    \caption{\textbf{Attention Circuit: Components \& Topology.}
    We ablate the attention-circuit design by varying grid mixing, identity embeddings, and aggregation.
    Row~A is our default configuration. 
    \textbf{Axial Conv} performs lightweight row/column mixing over the (Layer, Head) grid, offering a cheaper alternative to full global self-attention.
    }

    \label{tab:attn_ablation_components}
    \vspace{2mm}
    \centering
    \begin{tabular}{clcc}
    \toprule
    \textbf{ID} & \textbf{Configuration / Change} & \textbf{\#Params} & \textbf{AUROC} \\
    \midrule
    \rowcolor{blue!10} \textbf{A} & \textbf{Gnosis (Axial + PMA)} & \textbf{1.4M} & \textbf{0.92} \\
    \midrule
    \multicolumn{4}{l}{\textit{Grid Topology \& Identity}} \\
    B & Remove Axial Conv (Linear Proj Only) & 1.1M & 0.84 \\
    C & Replace Axial Conv w/ Global Transformer & 4.5M & 0.92 \\
    D & Remove Layer/Head Embeddings & 1.4M & 0.90 \\
    \midrule
    \multicolumn{4}{l}{\textit{Aggregation Strategy}} \\
    H & Replace PMA w/ Mean Pool (Axial $\to$ Mean) & 0.9M & 0.85 \\
    \bottomrule
    \end{tabular}
\end{minipage}
\hfill
% RIGHT TABLE: Hyperparameters (Layers & Downsampling)
\begin{minipage}[t]{0.38\linewidth}
    \caption{\textbf{Attention Circuit: Hyperparameters.}
    Impact of layer selection stride and initial map downsampling size ($k_{\text{grid}}$).
    (Proposed: Select 1 every 5 layers; $k_{\text{grid}}=256$). Note: These choices mainly affect inference speed/memory rather than parameter count. Higher AUROC is better.
}
    \label{tab:attn_ablation_hyperparams}
    \vspace{2mm}
    \centering
    \begin{tabular}{clc}
    \toprule
    \textbf{ID} & \textbf{Variation} & \textbf{AUROC} \\
    \midrule
    \rowcolor{blue!10} \textbf{A} & \textbf{Gnosis} & \textbf{0.92} \\
    \midrule
    \multicolumn{3}{l}{\textit{Layer Selection Strategy}} \\
    E1 & All Layers (Every Map) & 0.91 \\
    E2 & First \& Last Layers Only & 0.64 \\
    \midrule
    \multicolumn{3}{l}{\textit{Downsampling Size ($k_{\text{grid}}$)}} \\
    F1 & Small Grid ($64$) & 0.68 \\
    F2 & Large Grid ($512$) & 0.92 \\
    \bottomrule
    \end{tabular}
\end{minipage}
\end{table*}

\subsection{Hidden-State Circuit Encoder Ablations}
\label{app:ablation-hidden}

To identify the optimal architecture for the hidden-state circuit, we conducted
a comprehensive ablation study investigating feature dimensionality, local
temporal processing, and global aggregation strategies.
Table~\ref{tab:hidden-ablation-components} \& Table~\ref{tab:hidden-ablation-sizing} details this investigation.
The final design (\textbf{Row~A}) provides the best
trade-off between reliability estimation (AUROC) and computational efficiency.

\textbf{Dimensionality and Sizing.}
We first investigated the information bottleneck size ($d_{\text{tok}}, k_{\text{hid}}$).
Comparing Rows H and I against our proposed model (Row A) reveals a clear
performance plateau at size $192$.
Reducing the size to $96$ causes a sharp performance drop ($-0.04$ AUROC),
likely due to information loss during the initial pooling.
Conversely, scaling to $384$ triples the parameter count without any accuracy gain.
We therefore fix the dimensions to $192$ for efficiency.

\textbf{Local Temporal Encoder (Phase 1).}
We investigated whether explicit local feature extraction is necessary before
global processing.
Row~B shows that feeding raw pooled sequences directly to the set encoder reduces
performance by $0.04$ AUROC, confirming that Phase~1 acts as a critical
denoising stage.
Inside Phase~1, we found that architectural complexity matters: removing the
Squeeze-and-Excite gating (Row~C) or replacing the multi-scale dilated
convolutions with a standard convolution (Row~D) both degrade performance.
This suggests the model relies on capturing multi-scale temporal signals (via dilation)
and dynamic feature reweighting (via SE/Gating).

\textbf{Global Set Encoder (Phase 2).}
Finally, we analyzed the global aggregation stage.
We found that simply averaging the features (Row~F) or removing the global
self-attention refinement (Row~E) consistently hurts performance.
This validates the use of the SAB+PMA stack to capture global context and
learn specific reliability prototypes.
Notably, our hybrid design significantly outperforms a simple pooled MLP baseline (Row~G).

\begin{table*}[t]
\centering
\footnotesize
\setlength{\tabcolsep}{3pt}
\renewcommand{\arraystretch}{1.1}

% LEFT TABLE: Components and Baselines
\begin{minipage}[t]{0.58\linewidth}
    \caption{\textbf{Hidden-state Circuit: Components \& Baselines.}
    We ablate the hidden-state encoder by isolating the roles of local temporal mixing (Phase~1) and global set aggregation (Phase~2).
    Row~A is our default configuration.
    Removing Phase~1 (B--D) or weakening global aggregation (E--F) consistently reduces AUROC, and a simple pooled MLP baseline (G) performs substantially worse.
    }
    \label{tab:hidden-ablation-components}
    \vspace{2mm} % Space between caption and table
    \centering
    \begin{tabular}{clccc}
    \toprule
    \textbf{ID} & \textbf{Configuration / Change} & \textbf{\#Params} & \textbf{AUROC} & \textbf{$\Delta$} \\
    \midrule
    \rowcolor{blue!10} \textbf{A} & \textbf{Gnosis} & \textbf{2.6M} & \textbf{0.92} & \textbf{--} \\
    \midrule
    \multicolumn{5}{l}{\textit{Phase 1: Local Processing}} \\
    B & Remove Phase 1 (Raw Seq $\to$ Set Enc) & 2.4M & 0.89 & -0.03 \\
    C & Remove Gating \& SE (Sum only) & 2.6M & 0.91 & -0.01 \\
    D & Remove Multi-scale (Dilations $\to$ 1) & 2.4M & 0.90 & -0.02 \\
    \midrule
    \multicolumn{5}{l}{\textit{Phase 2: Global Aggregation}} \\
    E & Remove SAB (CNN $\to$ PMA only) & 1.3M & 0.85 & -0.07 \\
    F & Replace PMA w/ Mean Pool & 2.1M & 0.89 & -0.03 \\
    \midrule
    \multicolumn{5}{l}{\textit{Architectural Baselines}} \\
    G & Pooled MLP (GlobalPool $\to$ MLP) & 0.7M & 0.82 & -0.1 \\
    \bottomrule
    \end{tabular}
\end{minipage}
\hfill % Fills space between tables
% RIGHT TABLE: Sizing Investigation
\begin{minipage}[t]{0.39\linewidth}
    \caption{\textbf{Hidden-state Circuit: Hyperparameters.}
    Impact of varying Feature Dimension ($d_{\text{tok}}$), Pooled Sequence Length ($k_{\text{hid}}$), and SAB Depth.
    (Gnosis used settings: $d_{\text{tok}}=192, k_{\text{hid}}=192, \text{SAB}=3$)}
    \label{tab:hidden-ablation-sizing}
    \vspace{2mm}
    \centering
    \begin{tabular}{clcc}
    \toprule
    \textbf{ID} & \textbf{Variation} & \textbf{\#Params} & \textbf{AUROC} \\
    \midrule
    \rowcolor{blue!10} \textbf{A} & \textbf{Gnosis} & \textbf{2.6M} & \textbf{0.92} \\
    \midrule
    \multicolumn{4}{l}{\textit{Feature Dimension ($d_{\text{tok}}$)}} \\
    H1 & Small Width ($96$) & 0.8M & 0.88 \\
    H2 & Large Width ($384$) & 9.3M & 0.92 \\
    \midrule
    \multicolumn{4}{l}{\textit{Pooled Seq Length ($k_{\text{hid}}$)}} \\
    I1 & Short Seq ($96$) & 2.6M & 0.89 \\
    I2 & Long Seq ($384$) & 2.6M & 0.92 \\
    \midrule
    \multicolumn{4}{l}{\textit{SAB Layers (Default: 3)}} \\
    J1 & Fewer Layers ($1$) & 1.7M & 0.89 \\
    J2 & More Layers ($5$) & 3.5M & 0.93 \\
    \bottomrule
    \end{tabular}
\end{minipage}
\end{table*}

\section{Additional Experimental Setup}
\label{app:additional-exp-setup}

\subsection{Benchmarks (extended).}
We evaluate on three disjoint domains. For \textbf{math reasoning}, we use a combined test set of AMC12 2022, AMC12 2023\cite{aimo2024validationamc}, AIME 2024\cite{aime24}, AIME 2025\cite{aime25}, and HMMT February 2025\cite{balunovic_srimatharena_2025}. These competition-style problems span a wide range of difficulty and require multi-step symbolic and numeric reasoning. For \textbf{open-domain QA}, we use an 18k-question held-out trivia subset drawn from the same distribution as our training corpus but with no overlapping items. This benchmark emphasizes short factoid answers and directly evaluates hallucination detection and knowledge grounding. For \textbf{academic knowledge reasoning}, we use MMLU-Pro\cite{wang2024mmlu}, an out-of-distribution evaluation spanning 14 diverse domains (e.g., math, physics, law, psychology) that combines domain knowledge with multi-step reasoning, providing a broad test of generalization beyond the training mix.

\subsection{Metrics (extended).}
We frame correctness prediction as binary classification and report both ranking and calibration quality. AUROC measures how well a method ranks correct completions above incorrect ones (0.5 = chance, 1.0 = perfect). AUPR is reported with two complementary positive classes: \textbf{AUPR-c} treats \emph{correct} completions as positive and summarizes how well a method recovers correct answers with high precision across recall; \textbf{AUPR-e} treats \emph{incorrect} completions as positive and summarizes how well a method detects errors/hallucinations, which is often the more safety-relevant viewpoint under class imbalance. For probability quality, we report Brier Skill Score (BSS), where BSS $>0$ indicates improvement over a prevalence baseline, and Expected Calibration Error (ECE), where lower values indicate better alignment between predicted correctness probabilities and empirical accuracy.

\section{Backbone Outcome Statistics}
\label{app:backbone-outcomes}

To contextualize the results in Table~\ref{table1}, we report the \emph{raw outcome breakdown} of each frozen backbone on the three evaluation domains. While Table~\ref{table1} focuses on the \emph{quality of correctness prediction} (AUROC/AUPR/BSS/ECE) for different judges and internal methods, this table provides the underlying \textbf{base behavior} of each backbone: how often it is correct, how often it hallucinates, and how often it produces no valid final answer. 
\textbf{Importantly, non-response cases are reported here for completeness but are filtered out and not used in our correctness-prediction evaluation.} 
Thus, the metrics in Table~\ref{table1} are computed over the subset of generations with a valid, parsable answer. This table therefore serves two purposes: (i) it clarifies the intrinsic difficulty and failure modes of each backbone across domains, and (ii) it provides the context needed to interpret domain-dependent shifts in AUPR and calibration metrics reported in Table~\ref{table1}, which are more sensitive to the underlying prevalence of correct vs.\ incorrect \emph{answered} generations.

\begin{table*}[t]
\centering
% \small
\setlength{\tabcolsep}{6pt}
\renewcommand{\arraystretch}{1.05}
\caption{\textbf{Backbone outcome rates (\%) across evaluation domains.}
For each frozen backbone, we report the fraction of instances that are \textbf{correct} (Accuracy), \textbf{incorrect} (Hallucination), or \textbf{no-answer} (``I don't know''/refusal/empty).
No-answer cases are shown for completeness but are \textbf{filtered out} from our correctness-prediction evaluation (Table~\ref{table1}).}
\label{tab:backbone-outcome-stats}
\begin{tabular}{l l c c c}
\toprule
\textbf{Backbone} & \textbf{Domain} &
\makecell{\textbf{Backbone}\\\textbf{Accuracy}} &
\makecell{\textbf{Backbone}\\\textbf{Hallucination}} &
\makecell{\textbf{Backbone}\\\textbf{``I don't know''}} \\
\midrule

\textbf{Qwen3 1.7B-Hybrid} & Math-Reasoning & 44.87\% & 47.76\% & 7.37\% \\
\textbf{Qwen3 1.7B-Hybrid} & TriviaQA       & 33.54\% & 55.67\% & 10.79\% \\
\textbf{Qwen3 1.7B-Hybrid} & MMLU-Pro       & 66.09\% & 31.53\% & 2.39\% \\
\midrule

\textbf{Qwen3 4B-Thinking-2507} & Math-Reasoning & 63.27\% & 25.31\% & 11.42\% \\
\textbf{Qwen3 4B-Thinking-2507} & TriviaQA       & 53.73\% & 45.27\% & 0.99\% \\
\textbf{Qwen3 4B-Thinking-2507} & MMLU-Pro       & 72.45\% & 24.75\% & 2.79\% \\
\midrule

\textbf{Qwen3 4B-Instruct-2507} & Math-Reasoning & 57.51\% & 29.26\% & 13.22\% \\
\textbf{Qwen3 4B-Instruct-2507} & TriviaQA       & 56.89\% & 42.98\% & 0.13\% \\
\textbf{Qwen3 4B-Instruct-2507} & MMLU-Pro       & 71.69\% & 27.46\% & 0.85\% \\
\midrule

\textbf{Qwen3 8B-Hybrid} & Math-Reasoning & 49.42\% & 14.16\% & 36.42\% \\
\textbf{Qwen3 8B-Hybrid} & TriviaQA       & 62.22\% & 36.05\% & 1.74\% \\
\textbf{Qwen3 8B-Hybrid} & MMLU-Pro       & 71.93\% & 22.29\% & 5.77\% \\
\midrule

\textbf{OpenAI gpt-oss-20B} & Math-Reasoning & 52.53\% & 42.63\% & 4.84\% \\
\textbf{OpenAI gpt-oss-20B} & TriviaQA       & 63.70\% & 27.67\% & 8.62\% \\
\textbf{OpenAI gpt-oss-20B} & MMLU-Pro       & 68.74\% & 30.83\% & 0.44\% \\

\bottomrule
\end{tabular}
\end{table*}

\section{Gemini As judge.}
\label{app:gemini}
For the LLM-as-judge baseline, we use Gemini 2.5 pro as a \emph{generative reward model}. Given a question and a model-generated answer, the judge produces a brief correctness analysis and then emits a scalar score in $[0,1]$ at the end of the response. We parse the score from the required \texttt{<score>} tags. Evaluating Gemini via the API on the three benchmarks reported in Table~\ref{table1} cost approximately \$500 in total. Figure~\ref{box:gemini-system-prompt} shows the prompt for using Gemini as the judge.

% \begin{verbatim}
% SYSTEM_PROMPT: 
%     """
%     You are an objective, expert evaluator. 
%     Your task is to review a Question and a proposed Answer.
%     Determine if the Answer is correct, and accurate based on the Question.
%     Assign a correctness score between 0.0 (completely wrong) and 1.0 (perfectly correct).
    
%     IMPORTANT:
%     Output the final score inside <score> tags.
%     Example: <score>0.95</score> or <score>0.0</score>.
%     """
% \end{verbatim}

\begin{figure}[t!] % try top of page
\vspace{-0.25em}
\centering
\begin{minipage}{\linewidth}
\captionof{figure}{Gemini judge system prompt used for the LLM-as-judge baseline.}
\label{box:gemini-system-prompt}
\begin{tcolorbox}[
  title={SYSTEM\_PROMPT},
  colback=white,
  colframe=black!60,
  boxrule=0.6pt,
  arc=2pt,
  left=6pt,right=6pt,top=6pt,bottom=6pt,
  fonttitle=\bfseries,
]
\ttfamily
You are an objective, expert evaluator.
Your task is to review a Question and a proposed Answer.
Determine if the Answer is correct, and accurate based on the Question.
Assign a correctness score between 0.0 (completely wrong) and 1.0 (perfectly correct).

IMPORTANT:
Output the final score inside <score> tags.
Example: <score>0.95</score> or <score>0.0</score>.
\end{tcolorbox}
\end{minipage}
\vspace{-0.5em}
\end{figure}

\begin{figure*}[t]
    \centering
    \includegraphics[width=\textwidth]{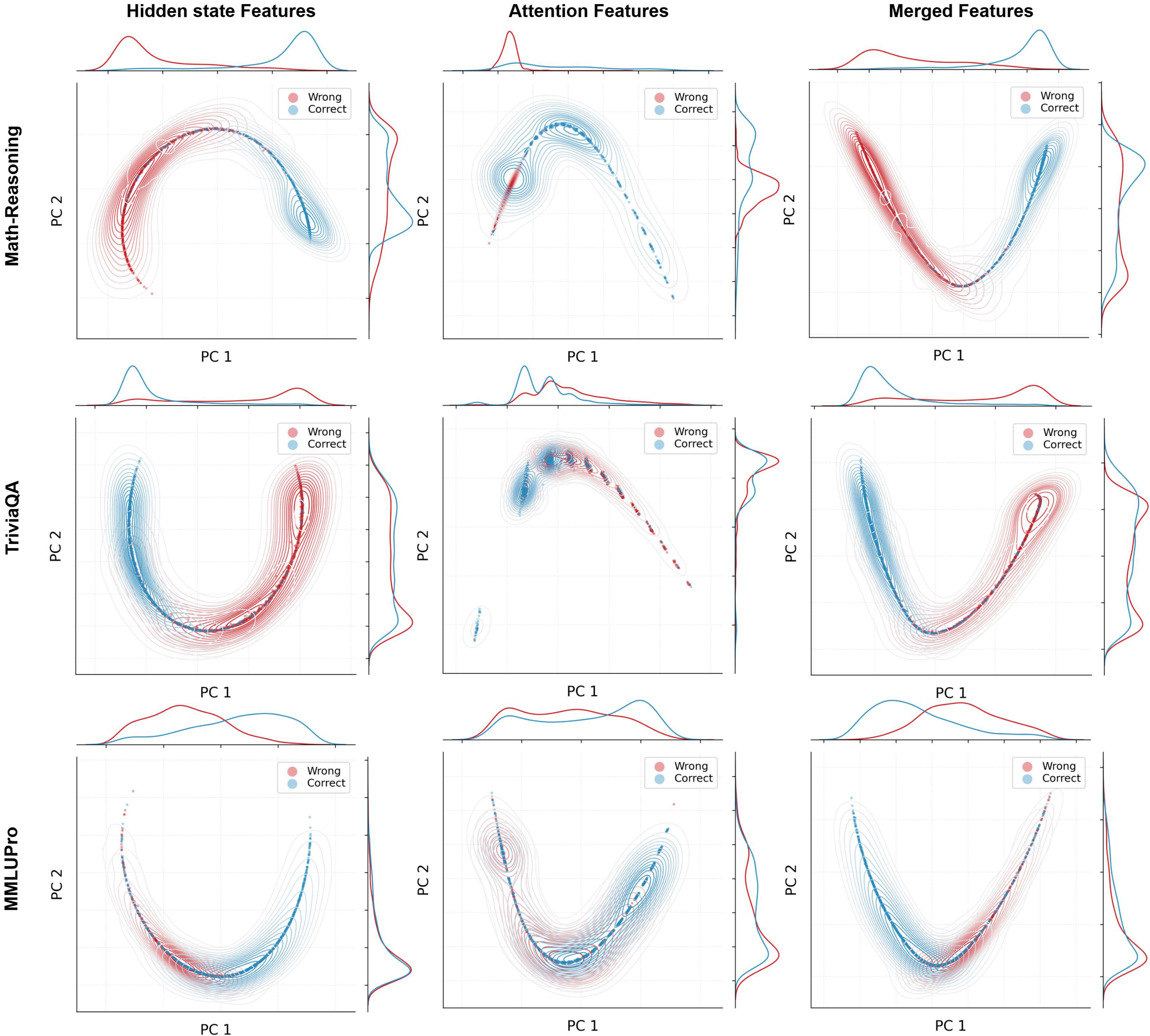}
    % \vspace{-2em}  % tighten space between image and caption
    \caption{
    2D PCA scatter plots of features learned by \textbf{Gnosis} across three domains.
    \textbf{Rows:} Math-Reasoning, TriviaQA, and MMLU-Pro.
    \textbf{Columns:} hidden-state features ($z_{\text{hid}}$), attention features ($z_{\text{attn}}$), and their merged representation.
    We show PCA scatter plots with KDE contours and marginal densities for \textbf{wrong} (red) and \textbf{correct} (blue) answers.
    Across domains, the merged features provide the clearest overall class separation, illustrating the complementarity of hidden and attention signals.
    }

    \label{fig:scatterplot_all}
\end{figure*}

\end{document}